\documentclass[manuscript]{acmart}

\setcopyright{acmlicensed}
\copyrightyear{2024}
\acmYear{2024}
\acmDOI{XXXXXXX.XXXXXXX}

\usepackage{xcolor}

\usepackage{multirow}
\usepackage{graphicx}
\usepackage{caption,booktabs}

\usepackage{subfig}
\begin{document}

\title{SleepNet: Attention-Enhanced Robust Sleep Prediction using Dynamic Social Networks}

\author{Maryam Khalid}
\affiliation{%
 \institution{Department of Electrical \& Computer Engineering, Rice University}
  \streetaddress{}
  \city{Houston, TX}
  \country{USA}}
\email{maryam.khalid@rice.edu}

\author{Elizabeth B. Klerman}
\affiliation{%
 \institution{Department of Neurology, Massachusetts General Hospital, Harvard Medical School}
  \streetaddress{}
  \city{Boston, MA}
  \country{USA}}
\email{ ebklerman@hms.harvard.edu}

\author{Andrew W. McHill}
\affiliation{%
 \institution{Sleep, Chronobiology, and Health Laboratory, School of Nursing, Oregon Institute of Occupational Health Sciences, Oregon Health \& Science University}
  \streetaddress{}
  \city{Portland, OR}
  \country{USA}}
\email{mchill@ohsu.edu}

\author{Andrew J. K. Phillips}
\affiliation{%
 \institution{Turner Institute for Brain and Mental Health, School of Psychological Sciences, Monash University}
  \streetaddress{}
  \city{Clayton VIC 3800}
  \country{Australia}}
\email{andrew.phillips@monash.edu}

\author{Akane Sano}
\affiliation{%
\institution{Department of Electrical \& Computer Engineering, Rice University}
\streetaddress{}
\city{Houston, TX}
\country{USA}}
\email{akane.sano@rice.edu}

\begin{abstract}

Sleep behavior significantly impacts health and acts as an indicator of physical and mental well-being. Monitoring and predicting sleep behavior with ubiquitous sensors may therefore assist in both sleep management and tracking of related health conditions. While sleep behavior depends on, and is reflected in the physiology of a person, it is also impacted by external factors such as digital media usage, social network contagion, and the surrounding weather. In this work, we propose SleepNet, a system that exploits social contagion in sleep behavior through graph networks and integrates it with physiological and phone data extracted from ubiquitous mobile and wearable devices for predicting next-day sleep labels about sleep duration. Our architecture overcomes the limitations of large-scale graphs containing connections irrelevant to sleep behavior by devising an attention mechanism. The extensive experimental evaluation highlights the improvement provided by incorporating social networks in the model. Additionally, we conduct robustness analysis to demonstrate the system's performance in real-life conditions. The outcomes affirm the stability of SleepNet against perturbations in input data. Further analyses emphasize the significance of network topology in prediction performance revealing that users with higher eigenvalue centrality are more vulnerable to data perturbations.

\end{abstract}

\ccsdesc[500]{Human-centered computing~mobile computing}
\ccsdesc[300]{Applied computing~Health care information systems.}

\keywords{social network, Graph convolution, graph neural networks, sleep, well-being prediction, contagion, wearable sensing, mobile computing, multimodal sensing }

\maketitle

\section{Introduction}
Sleep behavior is closely associated with physical ~\cite{sleep0,sleep01} and mental health~\cite{sleep1,sleep2}, both as an influence \cite{disease1} and as an effect \cite{cause}. Short sleep duration can increase the risk of multiple diseases and worsen existing conditions such as cardio-metabolic dysfunction \cite{disease1}, obesity \cite{disease2},  and diabetes \cite{disease3} and impact a person's functioning in daily life including impaired work efficiency \cite{sleepWork,sleepWork2}, fatigue, sleepiness, and performance deficits \cite{sleepWork3}.

To address these issues, mobile health (mHealth) applications are proposed for monitoring and providing appropriate interventions. The interventions that utilize behavior change techniques such as shaping knowledge, goals, and planning have been shown to have desirable effects \cite{sleep_reg_review} on sleep behavior and have great potential to mitigate the above-mentioned risks. These mHealth systems require accurate monitoring and prediction of sleep behavior. Sleep duration, when used in machine learning models as an input feature, can improve well-being prediction \cite{sleepSarah}. 
The prediction of sleep duration itself, however, can be a challenging task. While time-of-day and recent sleep/wake timing are strong predictors for the next sleep episode
duration, several external factors, such as social interactions and digital media usage, can also impact sleep duration \cite{media1,media2}.  The use of digital media can 
impact sleep duration directly or can lead to a person changing their behavior based on the people in their social network. There is a complex dynamic between an individual's social network, their digital media interactions encompassing calls, SMS messages, and app usage, and the influence these factors exert on that individual's behaviors. 

Previous studies have established the existence of sleep behavior "contagion" in a social network. Using a network of 8349 adolescents to investigate the relationship between sleep behavior and drug usage \cite{sleep-contagion}, Mednick and colleagues identified clusters of people with poor sleep behavior and demonstrated, through statistical analysis, how network centrality impacts sleep behavior outcomes. Additionally, these investigators found a connection between a person's future sleep duration and their friend's current sleep duration \cite{sleep-contagion}. 
Another aspect of the environment is weather which has also been shown to affect sleep duration \cite{weather3,weather4,weather5}. Weather impacts physical activity and sedentary behavior \cite{weather1,weather2}, which in turn impacts sleep behavior \cite{weather_sleep1,weather_sleep2}. Weather can also impact well-being and mood \cite{weather_mood}, both of which can affect sleep behavior and is, therefore, an important variable to consider when predicting sleep duration.

In this work, instead of estimating current sleep behavior, which is a well-studied topic in the published literature, we pose the problem of predicting the next night's sleep behavior from the current and past few days of data. Since short sleep duration is associated with poor performance and health conditions 
in people of all ages, we choose sleep duration as an attribute to quantify sleep behavior. Our work can theoretically be extended to the prediction of other attributes such as sleep timing or efficiency, or other health-related outcomes.

 \begin{figure}
     \centering
     \includegraphics[scale=0.52]{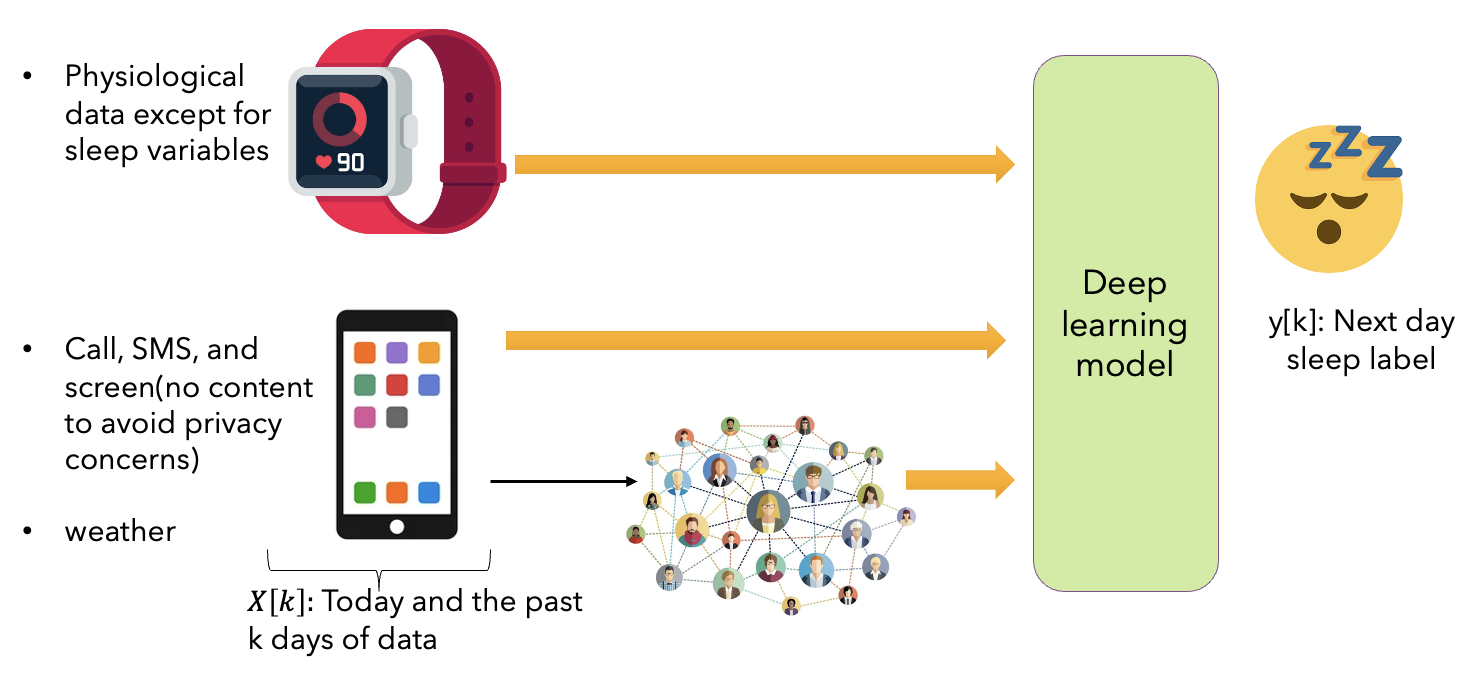}
     \caption{Framework overview: Multi-modal data from phone and wearable device is integrated with graph networks, extracted from phone data, in a deep learning architecture to predict the next day's sleep duration label}
     \label{fig:framework}
 \end{figure}
 
We exploit the relationship among social networks, digital media usage, and sleep duration in two ways. First, we use mobile phone data to capture digital media usage in the form of calls, SMS, and screen time (indicating app usage). Second, we incorporate the user's social network, based on calls and SMS usage, into the sleep prediction model as a graph architecture by integrating contagion through multi-user data aggregation. Our framework combines physiological data from wearable sensors, weather data, mobile phone usage, and social networks into a deep learning framework to predict next-day sleep duration labels (Fig.\ref{fig:framework}). Importantly, social networks extracted from mobile phone data do not incur additional user burden.

To this end, we use Graph Convolution Networks (GCN) to aggregate features from the user's neighbors and predict sleep duration labels for the entire network. However, irrelevant nodes (e.g., nodes that do not contribute to a user's sleep duration behavior) or noisy graphs can hinder prediction performance. To alleviate this problem, we incorporate an additional attention mechanism that filters out irrelevant nodes by paying more "attention" to important nodes by utilizing Graph Attention Networks (GAN). Another limitation of graph-based architecture is that it requires both a feature matrix and a graph network as inputs; this limits the size of graphs that can be handled by the model. However, in real life, the magnitude of social interactions often changes. To address this limitation, we utilize GEDD: Graph Extraction for Dynamic Graphs, an iterative algorithm based on the principle of graph convolution networks to transform a set of varying size graphs into a set of graphs with fixed predetermined size without any loss of the user or graph information. Lastly, we extract the temporal trends in the data through a Long Short-term Memory (LSTM) network module and integrate both multi-user aggregated features with extracted temporal trends in a deep learning architecture, abbreviated as SleepNet, that predicts the next-day sleep duration labels for all the users present in the network. 

We design multiple experiments to evaluate the proposed model and answer two important questions. First, we quantify the benefit of integrating social networks by comparing the performance of the proposed architecture to two benchmarks that do not utilize the social network, but otherwise are very similar to the proposed model in terms of the number of layers, units in each layer, hyperparameters, input, and output. The same samples from the dataset are used for training and testing all four models for a fair comparison. 
The evaluation results highlight a significant improvement provided by the integration of social networks. Between GCN and GAN, the attention-based model provides the highest accuracy. Second, we evaluate the impact of different design parameters on the performance of all models: specifically, the impact of network size and temporal memory is characterized through a set of experiments.

We further expand our analysis to investigate how the models perform in real-world scenarios where sensor data could be noisy or missing (e.g.,  the battery runs out or the sensor breaks down). Multi-user prediction models are more vulnerable because even if one user is producing low-quality data, the predictions of the entire network could be affected. To characterize this impact, we conduct a robustness analysis by perturbing the test data in different ways and observing the model's performance for the whole network as well as different parts of the network. Our results indicate that while GCN is prone to perturbations in the data, GAN is robust to these perturbations and has a negligible drop in performance except for nodes that have very large eigenvalue centrality.

To summarize, this work makes the following contributions,
\begin{itemize}
	\item  We present SleepNet, a framework that exploits contagion in sleep behavior for next-day sleep duration prediction. It utilizes rich multimodal data from wearables, mobile phones, surveys, and weather and integrates social networks in the prediction model. Furthermore, the architecture overcomes the limitations of noisy/missing-user graphs and large networks by leveraging an attention mechanism.

	\item We conduct a thorough experimental evaluation to quantify the improvement provided by the integration of social networks.

	\item We systematically analyze the model's robustness to data perturbations in real-world conditions. Our findings indicate that graph-based architecture is more robust to perturbations compared to non-graph-based architecture for multi-user predictions and that different parts of the network are impacted differently when a subset of the network is perturbed.

 \item We share a rich multi-modal time-series dataset about participants and their surroundings, including information from wearables, phones, surveys, and weather, collected from over 200 participants spanning  7500 days. Additionally, we provide the dynamic social network graphs for these participants constructed from call and SMS data representing diverse network topologies. The dataset is the first of its kind to integrate all four modalities with dynamic social networks in one longitudinal study.
	
\end{itemize}

\section{RELATED WORK}\label{relatedwork}

\subsection{Social contagion}
Contagion is "rapid communication of an influence (such as a doctrine or emotional state)" \cite{contagion_def}.  Contagion in behavior is the 
production of similar behavior in response to interaction with another person \cite{cont_def1,cont_def2,def3}. One example is the tendency to mimic and synchronize expressions, vocalizations, postures, and movements with those of another person. 
Multiple works have explored how health outcomes such as obesity \cite{cont_ref1},  tuberculosis 
 \cite{cont_ref2}, pneumonia \cite{cont_ref3}, and severe acute respiratory syndrome \cite{cont_ref4},  can spread in a network suggesting the existence of contagion in health behaviors \cite{cont_ref5}.
The work in \cite{sleep-contagion} suggests the contagion of sleep behavior in a social network.
An empirical study was conducted using data collected from students in grades
7–12 in the National Longitudinal Study of Adolescent Health \cite{NIHstudy} over 8 years. A social network of 8349 adolescents was extracted to study the relationship between sleep behavior and drug usage. The study found clusters of poor sleep behavior;  statistical analysis revealed that the probability of a person sleeping less than seven hours goes up by $11\%$ if their friend also sleeps less than seven hours. Further findings indicated that high network
centrality negatively impacts future sleep behavior. We extend this work to predict next-day sleep duration labels at a day-to-day resolution.

The literature closest to the proposed work was presented in \cite{network_MH} that leveraged social networks for health label prediction \cite{network_MH}. Characteristics about the position of a user in the weekly social networks extracted from smartphone data (SMS exchanged), such as centrality, were used as input features to a binary classifier to predict whether an individual was depressed/anxious or not. We expand this work by integrating rich physiological, smartphone, and weather data in the prediction model. Our architecture also significantly differs from this work: instead of using network metrics as model input, we \textit{actively} use the network to perform information aggregation from multiple users and exploit contagion at a deeper level.

\subsection{Sleep Sensing Technologies}

In this section, we discuss the sensing technologies used to extract sleep behavior information.

\subsubsection{Wearable Technologies}
The gold standard to measure sleep is polysomnography \cite{poly} which collects multiple physiological parameters, two of which are the electroencephalogram (EEG) and electrocardiogram (ECG). However, it is expensive and burdensome to participants and staff. A large body of literature investigates learning sleep characteristics from the information-rich brain and heart activity signals collected using EEG \cite{eeg3,eeg4} and ECG \cite{eeg5}. However, the collection of EEG and ECG data requires wearing electrodes at multiple locations of the body which is not convenient in real-life settings.

As an alternative to EEG and ECG data, wearable devices are easy to use, less obtrusive, portable, low-cost,  and can be worn without any difficulty when the individual chooses to go to bed/sleep.  The efficacy of using actigraphy sensors to predict sleep quality was investigated in \cite{wear2}. Learning sleep attributes from smartwatches is also thoroughly investigated in previous works \cite{wear3,wearable11}.  Multi-modal sensing \cite{ecg_wear1} extends this by combining data from the actigraphy sensor and ECG for understanding the time spent in different sleep stages. Our system is similar to most of these works, as we also utilize wearable acceleration sensors for predicting sleep duration. However, in addition to acceleration data, we incorporate the impact of skin temperature, skin conductance, surroundings, and digital media usage on sleep duration by utilizing weather and mobile metadata.

\subsubsection{Non-wearable Low User-burden Technologies}
Another interesting body of literature focuses on solutions in which the user does not have to wear any sensor at all.
For instance, a system that uses radio-signal transmitting devices in the room to monitor sleep biomarkers is non-invasive and low-burden \cite{zero_effort,dopple_sleep}. Another approach involves multiple sensors throughout a smart home to forecast sleep behavior \cite{noninvasive2}. However, these solutions are limited to controlled environments, can be costly, and are not effective when there are multiple people in the home or when the participant is not at home (e.g., traveling).

These limitations were overcome by work \cite{mobile_only,studentlife,tossnturn} in which only mobile phone data were utilized to learn sleep behaviors. However, these require the smartphone to be always placed close to the user so that features representing mobility, light, and activity can be extracted. Another alternative technology is a bed sensor e.g. EMFIT QS sleep analyzer, however, the current state of this technology does not provide performance comparable to wrist actigraphy \cite{mattress}.

In summary, there are two main types of sensing technologies based on proximity to the user. Close proximity sensors such as smartwatches and mobile phones are used by many people and are relatively low-cost. Distant proximity sensors such as radio transmitters and IoT (Internet-of-Things) devices require specialized setup.  Thus, there is a trade-off between user burden and sensor burden (e.g., cost, ubiquity) when choosing the sensing technology. In future work, insights from \cite{mobile_only,zero_effort,saeed_mobile_only} may be useful in reducing the burden of wearing the device.

\subsection{Modeling for Learning Sleep Attributes}

 \subsubsection{Estimation using Rule-based/heuristic Models} 
 In \cite{mobile_only}, mobile phone data and a 2-process sleep regulation model estimated sleep timing, showing improved performance with expert knowledge. In a related study \cite{saeed_mobile_only}, phone usage data inferred sleep duration using personalized rule-based algorithms. SleepGuard \cite{wear3} developed an end-to-end architecture for smartwatch data to estimate sleep stages, postures, movements, and sleep environment, utilizing heuristic-based models or a linear-chain conditional random field \cite{wear4}.

 \subsubsection{Estimation using Machine Learning Models}
 Machine learning models have been employed to overcome the limitations of rule-based approaches, which can lack generalizability to different populations and non-stationary dynamics of the environment.

 Linear regression models using smartphone-derived features related to mobility, light exposure, and usage were applied to infer sleep duration \cite{StudentlifeAlgo,StudentLifeAlgo2}. EEG and ECG data have been used to identify sleep disorders and classify sleep stages using various models, including k-nearest neighbor classifiers, convolutional neural networks (CNN) \cite{eeg2,eeg3,eeg4}, deep neural networks \cite{eeg5}, and support vector machines \cite{ecg2}. However, these models often relied on handcrafted features and may have limited generalizability.
 
Another work combined \cite{ecg_wear1}  actigraphy data with ECG data for calculating the duration of time spent in different sleep stages using 
multiple machine learning models.

In this paper, instead of monitoring or estimating current sleep behaviors, we aim to predict a sleep attribute for the next day; this is a more complex learning problem but with greater translational potential for sleep management and mHealth applications. To achieve this objective, we develop a deep learning architecture that predicts the next day's sleep duration for multiple users from wearable, mobile, and weather data.

\subsubsection{Prediction using Machine Learning Models}

Previous research has investigated the use of actigraphy sensors and various deep learning models like multilayer perceptron (MLP),  Recurrent Neural Network (RNN), Convolutional Neural Networks (CNN), Long Short-Term Memory (LSTM-RNN), and a time-batched version of LSTM-RNN (TB-LSTM) to predict sleep quality labels \cite{wear2} or next day's sleep duration \cite{wearable11}. Our approach preserves the sequential nature of daily features and trends by utilizing an LSTM network. In addition to wearable data, we incorporate the impact of surroundings and digital media usage on sleep by using weather and mobile data. Moreover, our system performs multi-user predictions while considering contagion effects between users in a graph-based deep learning architecture.

In all the above-mentioned works, feature extraction is critical. However, in real life, data sources can be noisy and data can be missing, which can impact the quality of extracted features. In order to investigate the performance of our system in the wild, we provide a thorough robustness analysis for the proposed model.

\section{Data Collection}\label{dataset}
Over a 5-year period from 2013 to 2017, we collected multi-modal data from 250 participants (158 female, 92 male) in the SNAPSHOT study \cite{sano2018} with an average age of 21 years (standard deviation: 1.3 years). The majority, about 240 participants, were aged 17-22, while 10 were between 22-28 years old.

 \textbf{Study Design and Participant Cohorts:}
 Each academic term represents a cohort, numbered chronologically from 1 (first) to 7 (last).  Each person participated for ~30 consecutive days (cohorts 1-6) or ~110 consecutive days (cohort 7) during one academic term. The numbers of participants in each cohort from 1-7 were 20, 48, 46, 47, 40, 35, and 15, respectively.  After preprocessing, over 80\% of the participants in cohorts 1-6 contributed data for 10-25 days, while in the last cohort, data were available for 40-65 days for around 75\% of the participants.

\textbf{Participant Eligibility Criteria:}
 All participants were students at one university. Participants were asked to identify 5 people whom they usually call or SMS at least once a week and all of them must meet the following criteria, (i) undergraduate students, (ii) own an Android smartphone, and (iii) age between 18 and 60 years. Exclusion criteria encompassed individuals with conditions such as wrist sensor discomfort (e.g., irritated skin), pregnant women, freshmen (cohort 1 and 3 only), and those who had traveled more than one time zone away one week before or had plans for such travel during the study (except for cohort 7). In the last two cohorts, friends of participants, even if not Android users, were included, emphasizing the recruitment of interconnected friend groups. We preferred recruiting bigger groups of people who were friends with each other.

 \textbf{Data Collection and Processing:}
 During the study period, the participants (i) wore a wrist device that collected skin conductance, skin temperature, and acceleration (Q-sensor, Affective, USA) on their dominant hand; (ii) had an app (funf \cite{funf}) installed on their phone that collected metadata of calls, SMS, screen usage, and location; (iii)  answered morning and evening daily surveys about drugs and alcohol intake, sleep time, naps, exercise, and academic and extracurricular activities. During these years of study, there were very few phone apps on which SMS or calls could be made, and therefore, our data collection about texting and calls made using the phone was considered to be complete. At the start and the end of the study, participants completed social network surveys including questions about social activity partners, emergency contacts, and roommates to identify people that participants would often interact with. For more details, please refer to Supplementary Material section  \ref{supp}. 

 Subsequently, windowed features were extracted from wearable device data containing multiple features for activity and physiology. Phone calls, SMS, and screen time were also summed over windows and used as features (e.g., number of events, timing, number of people communicated, distance traveled within a day or within different timeframes). Multiple features explaining weather conditions (e.g., temperature,  precipitation, cloud cover, pressure, humidity, sunlight, and wind) were added to the feature matrix. After removing all sleep-behavior labels, such as sleep efficiency, data from all modalities were concatenated to create one feature matrix with 317 features (details in Supplementary Material \ref{feat}).

 \textbf{Ethics Approval and Data Availability:} 
The study protocols and informed consent procedure were approved by the Massachusetts Institute of Technology and Partners HealthCare Institutional Review Boards. The study was registered on clinicaltrials.gov (NCT02846077). 
All participants signed an informed consent form. All methods were performed in accordance with the relevant guidelines. To protect study participants’ privacy and consent and since some of the participants did not consent to sharing their data with the third-party researchers, raw data will not be publicly available. 
However, deidentified data including the processed input features, labels, and graph networks extracted from phone data, which are used to evaluate the models in this work and would be needed for reproducibility, are available online \cite{datasetSpand11}.

\section{SleepNet: Sleep Prediction Architecture using Network Dynamics}

\subsection{Problem Description}
We pose a problem of predicting sleep labels for the next day based on wearable and mobile data for the current and past few days. Formally, the multi-modal data with $m$ features for a given day $k$ is represented by the vector $x[k]\in \mathbb{R}^{m} $ and the corresponding sleep duration in minutes is represented by $s[k]$. The feature matrix $X[k]$ is composed of past $L$ days of feature data, where $L$ is an indicator of temporal memory,  and the sleep label $y[k]$ is generated from sleep duration $s[k]$ (mins). The recommended sleep duration for young adults is between $8-10$ hours \cite{duration1, duration2, duration3}, so we choose a threshold of 8 hours for differentiating between good and poor sleep behavior. Also, the sleep duration distribution after preprocessing is relatively balanced around 8 hours.
The sleep duration for our collected dataset had a mean of 7.2 hours (std= 2.1 hours).

\begin{equation}\label{input}
	X[k] = \{x[k-L-1],x[k-L],...x[k-1]\}
\end{equation}

\begin{equation}
	 y[k] =
	\begin{cases}
		1 & \text{if s[k]>=480}\\
		0 & \text{otherwise}
	\end{cases}    
\end{equation}
The objective of this work is to develop a data-driven model $f(.)$ parameterized by $\theta$ that can predict sleep labels  from the  feature matrix $X$ and participants' social network represented by $\mathcal{G}$,
\begin{eqnarray}
\mathcal{L} =  \arg \min_{\theta} ||y-f(X,\mathcal{G},\theta)||_2
\end{eqnarray}

\subsection{Graph Network Construction}
We extract participants' social networks from mobile phone data to analyze their impact on sleep behavior. While an alternative could be using social media friend networks, these networks are not always indicative of real interaction. Participants may be connected to people they don't interact with, or to those who do not influence their behavior. Additionally, social media networks are larger and noisier, leading the model to aggregate irrelevant information and negatively impact prediction accuracy.

We validated social interactions among study participants using data from pre- and post-study social network surveys. These data allowed us to quantify the closeness between participants and create graph networks, which are detailed in Supplementary Material section \ref{supp-net}. Our findings affirm the presence of social interactions and phone communication among the study participants.

We represent a participant's social network information as a graph network \cite{networks}. Graph $\mathcal{G}$ is composed of two main components: nodes $\mathcal{V}$ and edges $\mathcal{E}$, 
 $\mathcal{G} = (\mathcal{V},\mathcal{E})$. The nodes, also known as vertices, are participants who were part of the study at a given time. Since the study was conducted in seven cohorts, we cluster all participants in each cohort as part of the same graph in the initial processing. The edges are connections between nodes and are indicative of how close two participants are. We model the edges as weighted links such that the weights are proportional to their interaction. For  convenience with mathematical operations, we utilize the matrix representation of a graph known as an adjacency matrix 
$\boldsymbol{A}$ where the value at $i^{th}$ row and $j^{th}$ column is represented by $A_{ij}$,

\begin{equation}
	A_{ij} =
	\begin{cases}
		w_{ij} & \text{an edge  $\mathcal{E}_{ij}$ exists from $\mathcal{V}_i$ \; to $\mathcal{V}_j$ }\\
		0 & \text{otherwise}
	\end{cases}       
\end{equation}
And $w_{ij}$ represents the weight of an edge between node $i$ and node $j$, where $i,j \in \{1,2,..,|\mathcal{V}| \}$.

We construct two graphs: the call graph $\mathcal{G}_c$, represented by adjacency matrix $\boldsymbol{A^c}$, and the SMS graph $\mathcal{G}_s$, represented by $\boldsymbol{A^s}$. We design them based on call and SMS metadata, respectively, collected over a time interval $[0, T]$. Representing an incoming call of duration $d$ seconds from participant $\mathcal{V}_i$ to $\mathcal{V}_j$ at time $t$ by $\mathcal{C}_{ij}[t]$ ,
\begin{equation*}
	\mathcal{C}_{ij}[t] = 
	\begin{cases}
		d & \text{$\mathcal{V}_i$ calls $\mathcal{V}_j$}\\
		0 & \text{otherwise}
	\end{cases}       
\end{equation*}

\begin{equation}
	A^c_{ij} =  \sum_{t=0}^{T}\mathcal{C}_{ij}[t]
\end{equation}

For text messages, we consider two types of incoming messages: normal SMS with a text message body (Class 1 message), and Flash SMS with no message body (Class 0 message). Denoting an SMS from participant $\mathcal{V}_i$ to $\mathcal{V}_j$ at time $t$ by $S_{ij}[t]$,

\begin{equation*}
	S_{ij}[t] = 
	\begin{cases}
		w_1 & \text{$\mathcal{V}_i$ sends $\mathcal{V}_j$ a Class 1 message}\\
		w_2 & \text{$\mathcal{V}_i$ sends $\mathcal{V}_j$ a Class 0 message}\\
		0 & \text{otherwise}
	\end{cases}       
\end{equation*}
Prioritizing Class 1 messages because of their stronger interaction $w_1 > w_2$, we construct the SMS graph,
\begin{equation}
	A^s_{ij} =  \sum_{t=0}^{T}S_{ij}[t]
\end{equation}
In the experimental evaluation, $w_1$ and $w_2$ were chosen to be 1 and $0.5$, respectively. Class 1 messages are given greater importance than Class 0 messages due to the presence of real conversations in Class 1 exchanges indicating a stronger interaction.
For each node in the adjacency matrix, there is associated feature data $X$ and label data $y$. Only phone data is used to construct these networks; pre and post-study surveys do not contribute to these graphs.

\subsection{SleepNet: Proposed Architecture}

We develop a system that integrates information from multiple participants in a social network such that the contagion in sleep behavior can be utilized to better predict the next day's sleep duration.

\subsubsection{Multi-user Context Aggregation}
\begin{figure}[h!]
	\includegraphics[scale=0.37]{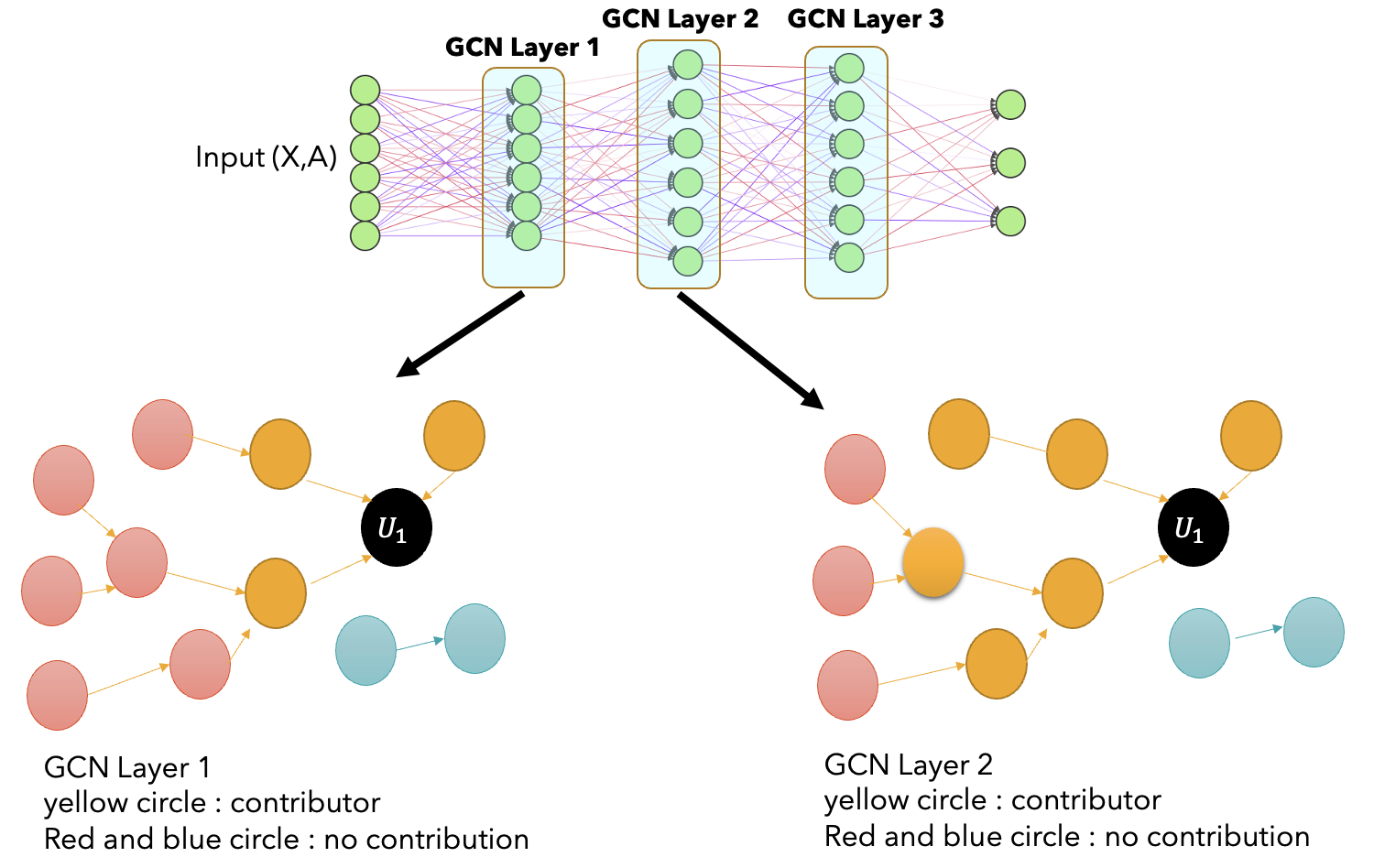}
	\caption{Understanding the principle behind deep GCN architecture. The top figure represents a multi-layer GCN network. The bottom figures show how information aggregation is performed in each layer for a given user $U_1$.}
	\label{GCN-1}
\end{figure}

In order to learn from both participants' data and their social networks, simple deep learning architecture such as neural networks cannot be used. The adjacency matrix representing the graph is permutation-invariant; i.e., the relative placement of nodes in the matrix is irrelevant. 
However, a simple neural network that uses the adjacency matrix as input would consider all permutations of the same matrix to be different inputs. Such an architecture cannot utilize graph connections between friends to aggregate information from the feature matrix in a systematic way such that the model accounts for both the user's and their friend's data when making a prediction.

To overcome these issues, we leverage graph convolution networks (GCN), which are inspired by spectral convolution on graphs \cite{GCN}.  GCN provides a layer-wise linear propagation rule that allows the network to learn from graphs. Spectral graph convolution is the convolution of any signal $x$ with a filter $g$, where the filter $g$ is derived from the graph. In order to compute spectral convolution, we need the laplacian and degree matrix of a graph. The degree matrix $D$ is a diagonal matrix containing degrees of the nodes on the diagonal where the degree of a node is a sum of incident edges. 
More details on the GCN layer can be found  in \cite{GCN}. 
The input to the first layer is the node feature matrix $X=[x_1,..x_N]$ containing data for all users in a graph, where $x_i \in \mathbb{R}^{m}$ is the feature vector for user $i$ and $N$ is graph size.

To predict sleep labels using this model, labels for each participant/node are used to compute the loss, and the model parameters are learned through forward and backpropagation. The training process and parameter tuning are similar to conventional neural networks. The only difference is that in addition to a feature matrix $X$, a graph adjacency matrix $A$  computed through call and SMS interaction data is also used as input to the model.

A deep model consists of multiple GCN layers stacked in a sequence. The first layer aggregates information from first hop neighbors, as shown in Fig. \ref{GCN-1}. Consider the participant $U_1$ shown in a black circular node whose multi-hop neighbors are represented by yellow and red nodes. Blue nodes represent participants who are not connected to $U_1$ at all. The bottom left figure represents the working principle on layer 1. In layer one, information from only first-hop neighbors in yellow is aggregated and passed on to the next layer. In the second layer shown on the bottom right, friends of friends of $U_1$ or its second-hop neighbors also contribute. The third layer, not expanded in the figure, also extends the same principle and aggregates information from third-hop neighbors. The blue nodes that are not connected to $U_1$ never contribute no matter how deep the network is. To summarize, an $\mathcal{H}^{th}$ hidden GCN layer aggregates features from up to $\mathcal{H}$-hop neighbors.

A key issue with this architecture is that the aggregation is solely based on edge characteristics. When the edges in constructed graphs are noisy or are constructed from heuristics that do not capture contagion in the sleep labels being investigated, information aggregation can actually negatively impact the performance. For instance, graph networks created from social media activity can contain connections with people who do not actually impact a participant's well-being. Thus, architecture learning solely on edge characteristics is not robust to noise. Even when there is no noise, dense graphs can contain irrelevant connections that harm the prediction performance. While there is proof for the existence of contagion, details are not known. Some people can mimic the sleep behavior of their friends and some do not, and the reasons are not known. A well-defined objective model for contagion is not available either. Thus, it is important to consider other factors that reflect the similarity between two participants when aggregating information.

\subsubsection{Attention-based Context Aggregation}
To overcome the above-mentioned issues, we modify the aggregation process and deploy an attention mechanism to identify \textit{important} contributing participants. To achieve this goal,  we incorporate graph attention networks (GAN). GAN considers two factors when combining information from multiple participants: graph edges and neighbor importance based on features. For a given participant $u_i$, GAN first considers all its neighbors as indicated by the adjacency matrix. In the second step, it assigns a significance coefficient to each neighbor that reflects the contribution that the node has in prediction for $u_i$. 

GAN performs information aggregation in two steps. For a graph of size $N$ with nodes $u_i$, $i\in\{1,2,..,N\}$,  the input to first layer is a stacked feature matrix $X = [x_1,x_2,...,x_N]$, where $x_i\in \mathbb{R}^{m} $ represent node $u_i$'s feature vector. The first step applies a learnable  linear transformation $W\in \mathbb{R}^{\tilde{l} \times m}$ to a feature matrix to obtain a high-level representation of the feature space,
\begin{equation}
	WX = [Wx_1,Wx_2,....,Wx_N]=[\tilde{x}_1,\tilde{x}_2,....,\tilde{x}_N]
\end{equation}
In the second step, a significance coefficient $e_{ij}$, which represents how important node $u_j$ is to $u_i$, is computed. A shared attention mechanism $a : \mathbb{R}^{m}  \times \mathbb{R}^{m}  \rightarrow \mathbb{R} $ is applied to the transformed feature vectors \cite{GAN},
\begin{equation}
	e_{ij} = a(\tilde{x}_i, \tilde{x}_j)
\end{equation}
The attention mechanism used in this work applies a non-linear activation $\sigma$ to a product of a trainable kernel $g \in \mathbb{R}^{2m}$ and concatenated feature vectors,
\begin{equation}
	e_{ij} = \sigma (g[x_i \ensuremath{+\!\!\!\!+\,}  x_j])
\end{equation} 
where $ \ensuremath{+\!\!\!\!+\,} $  represents the concatenation operator.

To integrate graph structure in this attention mechanism, these coefficients are only computed for the first-hop neighbors. In addition, to compare different neighbors and have stable information aggregation, the coefficients are normalized across all neighbors. 
Finally, the normalized coefficients are used to compute a weighted aggregation of features from all over the node $u_i$'s neighbors.

The attention mechanism mitigates the impact of non-contributing nodes in dense graphs and is robust to noisy edges. 
This architecture also offers higher generalization to graph structures not seen by the model during the training process.

\subsubsection{Dynamic User Distribution in Graphs}\label{GEDD}

In our dataset, the size of the extracted social networks varied as the number of participants in each cohort, and their connections were different. The variation in the number of participants per cohort is challenging because the proposed graph-based models require both features and graph networks as input. Both these inputs \textit{fix} the size of the input layer. When the number of participants changes, the size of the graph deviates from the model's pre-determined input size. 

To overcome this problem of varying network sizes, we use an algorithm called Graph Extraction for Dynamic Distribution (GEDD) \cite{gedd}. GEDD is a connected component-based method that converts large dynamic graphs into a set of small graphs of size equal to the model's input size  $\eta$. GCN performs feature aggregation from up to k-hop neighbors in the $k^{th}$ layer and no aggregation from users that are neither direct nor indirect neighbors. GEDD exploits this concept for extracting graphs of size $\eta$ through connected components. A connected component of a graph is a subgraph in which each node is connected to another through a path \cite{graph}. For a graph with $\tilde{N}$ nodes $	\mathcal{G_{\tilde{N}}=(V,E) }$, there are $\omega$ connected components with $1\leq \omega\leq \tilde{N}$. When $\omega=1$, all the nodes in $\mathcal{G_{\tilde{N}}}$ are connected, and when $\omega=\tilde{N}$, all nodes are disconnected and have a zero degree. The breakdown of graphs in connected components will result in subgraphs of varying sizes. Let $\Phi_i$ 
represent the $i^{th}$ connected component,
\begin{equation*}
	\Phi_i = \{\mathcal{V}^j,\mathcal{E}^j\} \;\; i={1,2,..\omega};\; j={1,2,,...\tilde{N}}
\end{equation*}
and $|\Phi_i|=q_i, 1\leq q_i \leq \tilde{N}$ represent the size of the component. 
First, the components are divided into two containers, Main container $\mathbb{D}$ and residue container $\mathbb{F}$, based on their size. The former will contain subgraphs of size $\eta$ and the residual will contain graphs of size $r< \eta$. This leads to three scenarios,
\begin{itemize}
	\item when $q_i=\eta$, add $\Phi_i$ to  $\mathbb{D}$
	\item when $q_i<\eta$, add $\Phi_i$ to  $\mathbb{F}$
	\item when $q_i>\eta$, break $\Phi_i$ into $\mathcal{Q}=\lceil \frac{q_i}{\eta}  \rceil$ subgraphs $\Phi_i^b$ where $b={1,2,..\mathcal{Q}}$. The large component is broken  such that,
	\begin{equation*}
		\Phi_i^b =
		\begin{cases}
			\eta & \text{ $b={1,2,..\mathcal{Q}-1}$}\\
			q_i \bmod \eta & \text{$b=\mathcal{Q}$}
		\end{cases}       
	\end{equation*}
	The subgraphs that satisfy the first condition in the above equation $\{\Phi_i^1,\Phi_i^2,...,\Phi_i^{\mathcal{Q}-1}\}$ are added to $\mathbb{D}$  and $\Phi_i^\mathcal{Q}$ to $\mathbb{F}$.
\end{itemize}
Once the components are divided between two containers, the main container is ready to be fed to the model. During the process of creating smaller graphs, stronger ties are given preference, and the graph is broken at weaker ties. For the residue container with all subgraphs smaller than $\eta$, the algorithm concatenates multiple subgraphs to create size $\eta$ subgraphs. There is still some residue left at the end when the \textit{total} number of nodes in $\mathbb{F}$ are less than $\eta$. For this last set, the algorithm uses repetition of nodes to create a final size $\eta$ subgraph. 

\subsubsection{Temporal Dynamics}
While graph-based models proposed in earlier sections are able to learn from multiple participants, the data collected from wearable and mobile phones also contain latent temporal trends for which the data need to be analyzed over a period of time. When utilizing multi-modal data for sleep behavior prediction, it is important to realize that instantaneous values of many sleep-related indicators might not be very informative on their own. Several factors lead to a certain state. For example,  today's sleep duration may not be explained by data collected during that day but instead from the temporal dynamics of the same features for the past few days. To integrate these dynamics into the model, we develop a spatiotemporal model, that captures multi-participant spatial domain through graph-based networks and for the temporal domain, we employ long-short term memory (LSTM) networks \cite{lstm}, with recurrently connected memory modules.

For the sleep duration label prediction problem, we extract the sequential information in features to predict the sleep label $y$. For a given participant, let $x[k]\in \mathbb{R}^n$ and $y[k]$ represent the stacked feature vector and sleep label for day $k$ respectively for all users.  If $L\in  \mathbb{Z}^+$ is the length of the sequence, then we create an $n\times L$ sequence matrix,
\begin{equation}
	S_k^L =\big[x[k-L]] ,x[k-L+1],...x[k]\big]
\end{equation}
This sequence matrix serves as an input feature matrix for which we predict the future sleep label $y[k+l]$. Here $l \in  \mathbb{Z}^+$ represents how far in the future we want to make the prediction. The tuples $(S_k^L,y[k+l])$ are used to train an LSTM network in a supervised fashion. The model weights are learned through conventional forward and backward propagation through the model with gradient descent.

\subsubsection{End-to-End Model}
\begin{figure}
    \centering
    \includegraphics[width=0.9\textwidth]{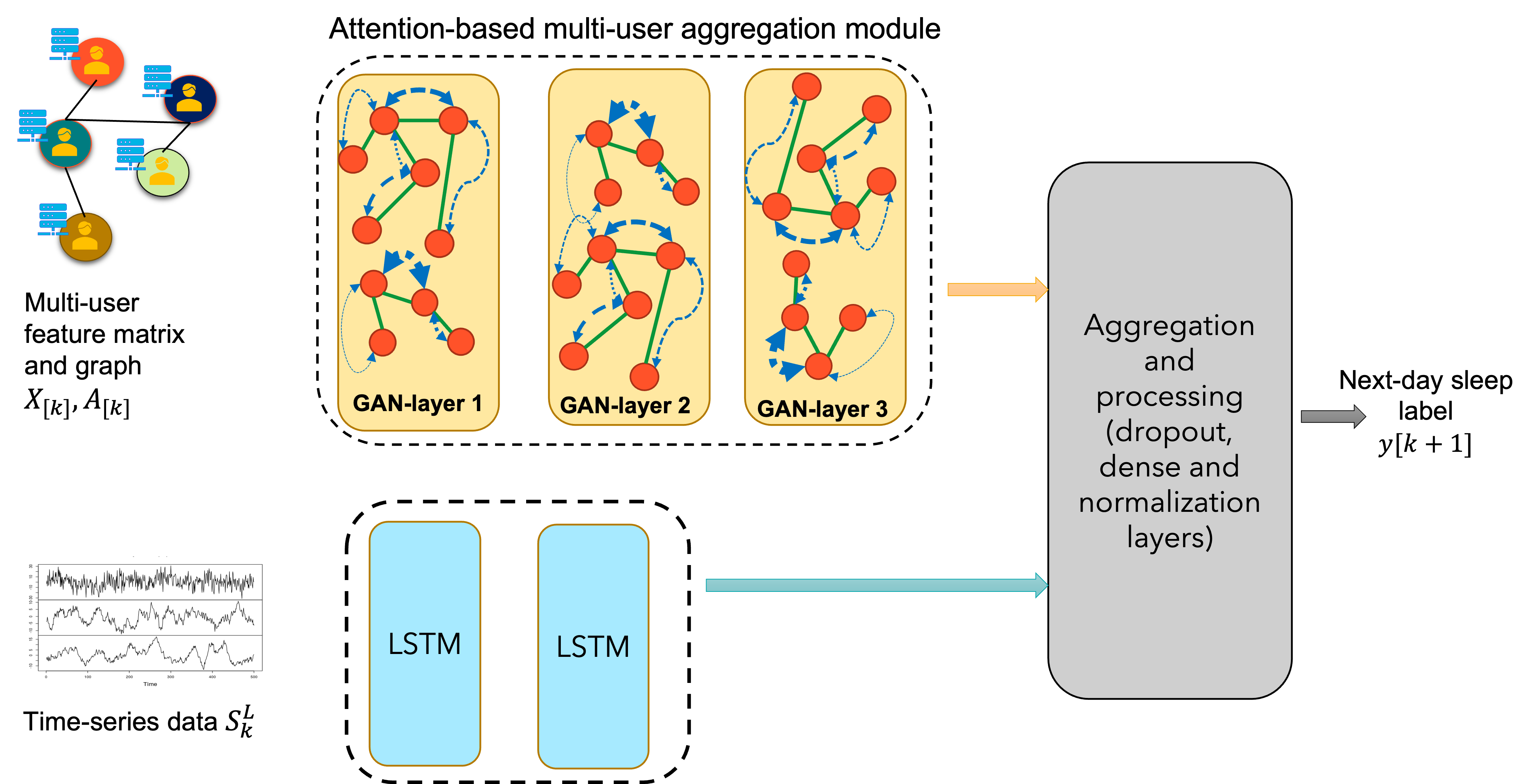}
    \caption{SleepNet Architecture. The top module based on GAN takes both the feature matrix and graph as input. The layer operates on graphs by applying attention shown by blue dashed links and original graph links are shown in solid green lines. The bottom module takes sequential data as input and extracts temporal trends. Finally, embedded features from both modules are combined in the aggregation module that produces predicted sleep labels.}
    \label{fig:arch}
\end{figure}
The complete architecture of SleepNet is shown in Fig.\ref{fig:arch}. 
The input to the model is a 3-tuple object $I = (X[k],S^L_k,A_N^k)$, where $X[k]=   \left[ x_1[k],x_2[k],...x_N[k]    \right]$ is a matrix containing feature vectors of $N$ participants each represented by $x_i \in \mathbb{R}^m$ for day indexed by $k$. The sequence data fed to the LSTM unit is represented by $S^L_k$. The graph network corresponding to the $N$ participants in feature matrix $X[k]$ is represented by adjacency matrix $A_N^k$. This corresponding data label is sleep duration for all $N$ participants for the next day represented by $y[k+1]$.
After extracting temporal dynamics and multi-participant aggregated features in an embedded space, the output of both modules is normalized and concatenated before being fed to a small network of dense layers that produce the final output. 

\subsection{Performance Evaluation}
Multiple experiments are conducted to evaluate different aspects of the proposed model. Firstly,  we quantify the improvement in prediction, if any, provided by aggregating information from multiple participants of a social network. We further investigate the impact of different network sizes and temporal memory on prediction performance. The preliminary evaluation with both SMS and Call graphs did not find much difference in the performance.  However, we prefer to report the result for SMS graphs because they are more dense, thus containing a wider distribution of network topologies (and therefore centralities). This is helpful in obtaining insights into how the user's position in the network impacted the model's robustness.

\subsubsection{Performance Comparison Models}
It is important to benchmark the proposed method against models that do not integrate the social network in the prediction of sleep duration behavior to quantify the impact of network integration. To achieve this goal, we consider two baseline models:
\begin{itemize}
	
		\item \textbf{CONV-LSTM: } As mentioned in section \ref{relatedwork}, multiple works use convolutional neural networks in sleep behavior estimation models from EEG and wearable data. Also, in recent work \cite{han2}, CNN was utilized to aggregate information from multiple modalities for well-being prediction. Thus, we utilize CNN to learn the multi-participant spatial information and LSTM for temporal dynamics,  followed by dense and dropout layers. This architecture is identical to the proposed method except that the graph attention network is replaced by CNN.

	\item \textbf{LSTM only:}  In order to observe the improvement provided by graph integration, we also evaluate the model with LSTM layers \cite{lstm_ref} only and no GAN. Similar to the proposed model, the LSTM layer is followed by multiple dense and dropout layers.
	
\end{itemize}

\subsubsection{Machine Learning Pipeline}\label{ml-pipe}
There are three main steps performed in preprocessing the data: processing missing data, removing the outliers, and standardizing the multimodal data. The details of preprocessing are provided in Supplementary Material section \ref{preprocess}.
After preprocessing the feature data and labels, the data are aligned with the corresponding graphs as each sample of $N$ participants has an associated 3-tuple: $(X[k], S^L_k, A_N^k)$. In this tuple, $k$ is an index for time(day), $L$ represents temporal memory and $N$ represents the number of users in the network. 

The data for all the users across the cohorts are combined in one batch where each sample of the batch is composed of a graph network for a given day, the corresponding daily data feature vector for all users, and the sequential data for the past $L$ days for all those users. Once the batch of 3-tuple data samples is created, it is split into training and test sets in two ways:
\begin{itemize}
    \item \textbf{Random Split: } The dataset is randomly split into training and test sets. We use random splitting to avoid selection bias. Around $50\%$ of data is used for training, $10\%$ for validation, and the remaining $40\%$ for testing with SMS graphs. The same training data is used to train all 4 models and then the same test data is used for testing across all models. This ensures that a fair comparison is conducted between the models. In order to mitigate the impact of model sensitivity to parameter initialization and depict the stability of the model, we use bootstrapping \cite{bootstrap2} and repeat the training and testing procedure 25 times and report average results along with the standard deviation across trials.

    \item \textbf{Leave-one-cohort-out (LOCO): } Random split introduces a chance of data from the same subjects appearing in both training and test set. To conduct a subject-dependent evaluation, we randomly choose one cohort as a test set and use the remaining cohorts for training and validation sets. Again the same training data is used to train all 4 models and then the same test data is used for testing across all models. For bootstrapping this process, the process is repeated six times without test set replacement i.e. the test set cohort is different in all trials. Since subjects in all cohorts are different, this testing is also equivalent to leave-subjects-out cross-validation.
    
\end{itemize}

To reduce the computation time, we use early stopping  with a patience value of 30 epochs and minimum increment $\delta =0.01$. This ensures that if validation loss is not changing for more than 30 epochs by more than $\delta$, the training process stops. We utilize an ADAM optimizer with the same learning rate (0.001) for all three models. 

\subsubsection{Evaluation Metrics}
The output of the model is the sleep label which is a binary variable for all the participants in the graph.
We report the accuracy score (0-1), which is computed by dividing the total number of correct predictions by total predictions. To establish statistical significance between different model's performance, we conduct an Analysis of Variance (ANOVA) test \mbox{\cite{anova}} and Kruskal–Wallis non parametric test\cite{kWallis}. These tests compare the performances of the four models across all trials and test the null hypothesis that there is no statistically significant difference between the performance of different models. If the p-value is less than $\alpha=0.05$, we reject the null hypothesis. For further pair-wise comparison between proposed models and benchmarks, we conduct Tukey HSD (honestly significant difference) test\mbox{ \cite{tukeyHSD} }for each pair between the proposed model and benchmarks.

\section{Results for Model Evaluation}

\subsection{Sleep Duration Prediction}

\begin{table}[ht]
\centering
\begin{minipage}[t]{0.5\linewidth}
\centering
\resizebox{\columnwidth}{!}{%
\begin{tabular}{|c|cc|}
\hline
\multirow{2}{*}{\textbf{Model}} & \multicolumn{2}{c|}{\textbf{Mean Accuracy Score (std)}} \\ \cline{2-3} 
 & \multicolumn{1}{c|}{\textbf{Random Split}} & \textbf{LOCO} \\ \hline
LSTM & \multicolumn{1}{c|}{0.75 (0.01)} & 0.71 (0.03) \\ \hline
CONV-LSTM & \multicolumn{1}{c|}{0.76 (0.05)} & 0.77 (0.02) \\ \hline
Proposed GCN-LSTM & \multicolumn{1}{c|}{0.81 (0.02)} & 0.83 (0.02) \\ \hline
Proposed GAN-LSTM & \multicolumn{1}{c|}{0.83 (0.02)} & 0.84 (0.02) \\ \hline
\end{tabular}%
}
\caption{Evaluation Results.} 
\label{tab:eval}
\end{minipage}%
\hfill
\begin{minipage}[t]{0.42\linewidth}
\centering
\small 
\resizebox{\columnwidth}{!}{%
\begin{tabular}{|c|c|c|c|}
\hline
\textbf{Model} & \textbf{Benchmark} & \textbf{\begin{tabular}[c]{@{}c@{}}p-value \\ (Random \\ Split)\end{tabular}} & \textbf{\begin{tabular}[c]{@{}c@{}}p-value\\  (LOCO)\end{tabular}} \\ \hline
GCN-LSTM & LSTM &  $1.0x10^{-4}$ & 0.00 \\ \hline
GCN-LSTM & CONV-LSTM &  $2.0x10^{-4}$ &  $1.0x10^{-3}$ \\ \hline
GAN-LSTM & LSTM & 0.00 & 0.00 \\ \hline
GAN-LSTM & CONV-LSTM & 0.00 & 0.00 \\ \hline
\end{tabular}%
}
\caption{Tukey-HSD adjusted p-values.\\}
\label{tab:tukey}
\end{minipage}
\end{table}

The performance over a randomly chosen test set is evaluated in terms of accuracy score and the process is repeated 25 times for each graph size and temporal memory. A similar procedure is repeated for Leave-One-Cohort-Out (LOCO).  The mean accuracy score and the standard deviation in accuracy across the trials are reported in Table \ref{tab:eval}. ANOVA reveals that differences between models' means for both types of the split are statistically significant (Random split p-value$=5.2x10^{-8}$, LOCO p-value$=6.3x10^{-15}$).
 Additionally, the Kruskal-Wallis non-parametric test also confirms that there is a significant difference among different models (Random split p-value$=3.2x10^{-6}$, LOCO p-value$=4.5x10^{-9}$).

 LSTM alone performs the worst among all models. Adding CNN slightly improves the performance because it uses the kernel to observe and slide across chunks of data from multiple participants. However, since there is no systematic mechanism for aggregating information from multiple participants, the improvement is very small. CONV-LSTM is also the most unstable among all given the high variation in the accuracy score. The proposed method GCN-LSTM incorporates additional information about inter-participant relationships and therefore the performance is improved significantly. Our final model, GAN-LSTM further incorporates attention while aggregating information from multiple participants resulting in the best performance. The variance in the performance of the proposed models is also small. ANOVA confirms that the difference in the performance of all four models is statistically significant. Further pair-wise comparison between proposed models and benchmarks through Tukey HSD results in all adjusted p-values$< \alpha=0.05$ as shown in Table \ref{tab:tukey}. This establishes that the proposed models perform statistically significantly better than the benchmarks. 

\subsection{Impact of Network Characteristics}
The average performance reported in the previous subsection masks the impact of varying network sizes. A larger graph with dense edges is expected to have larger information aggregation than a smaller graph and that can significantly impact the accuracy of predictions. Thus, to characterize this impact, we fix temporal memory and vary graph size, evaluating the model in a random test-train split loop. 

\begin{figure}
	\centering
	\subfloat[][\centering Impact of graph size with short sequence]{{\includegraphics[width=6.7cm]{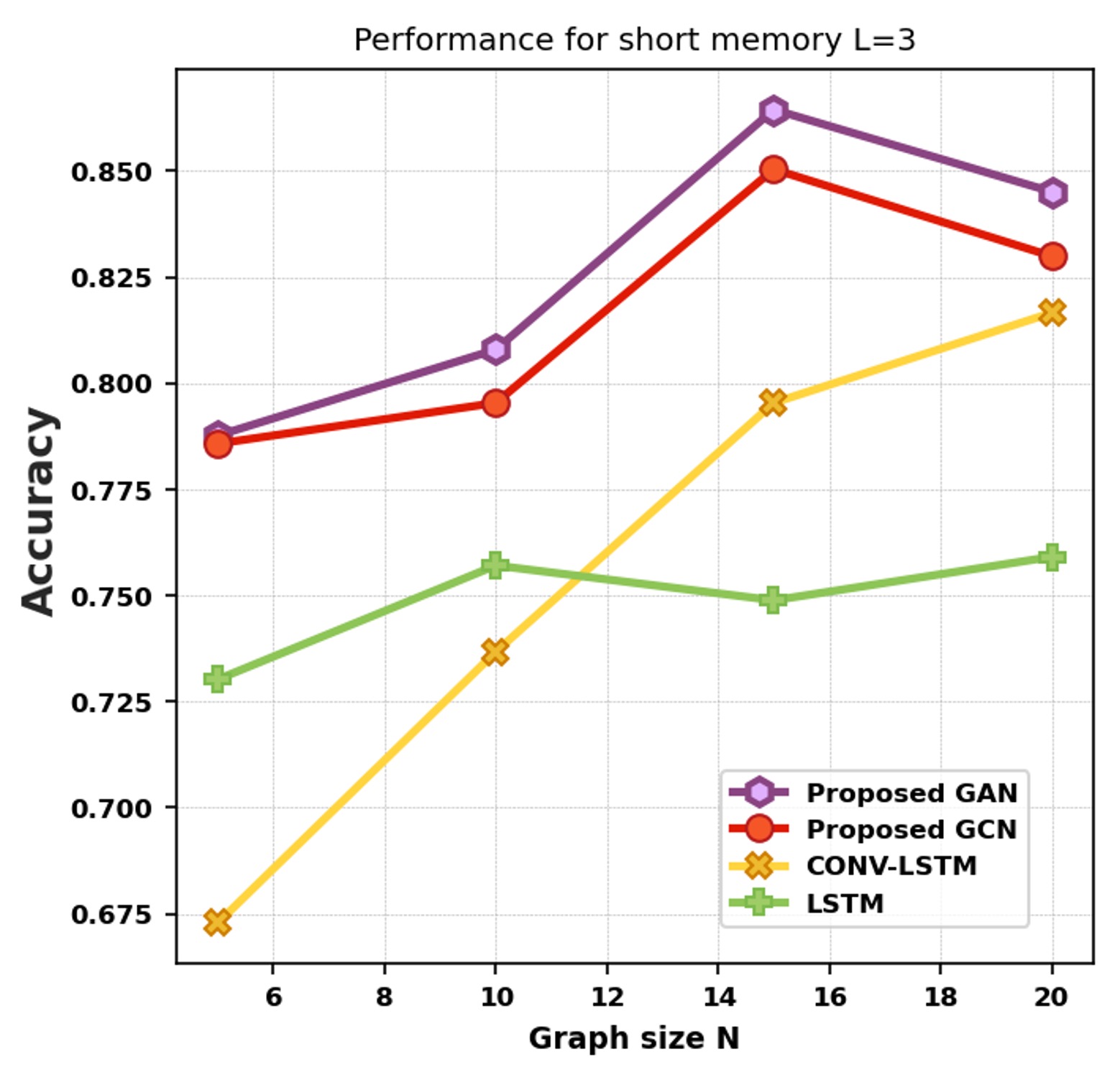} }}%
	\qquad
	\subfloat[][\centering Impact of temporal memory with small network]{{\includegraphics[width=6.7cm]{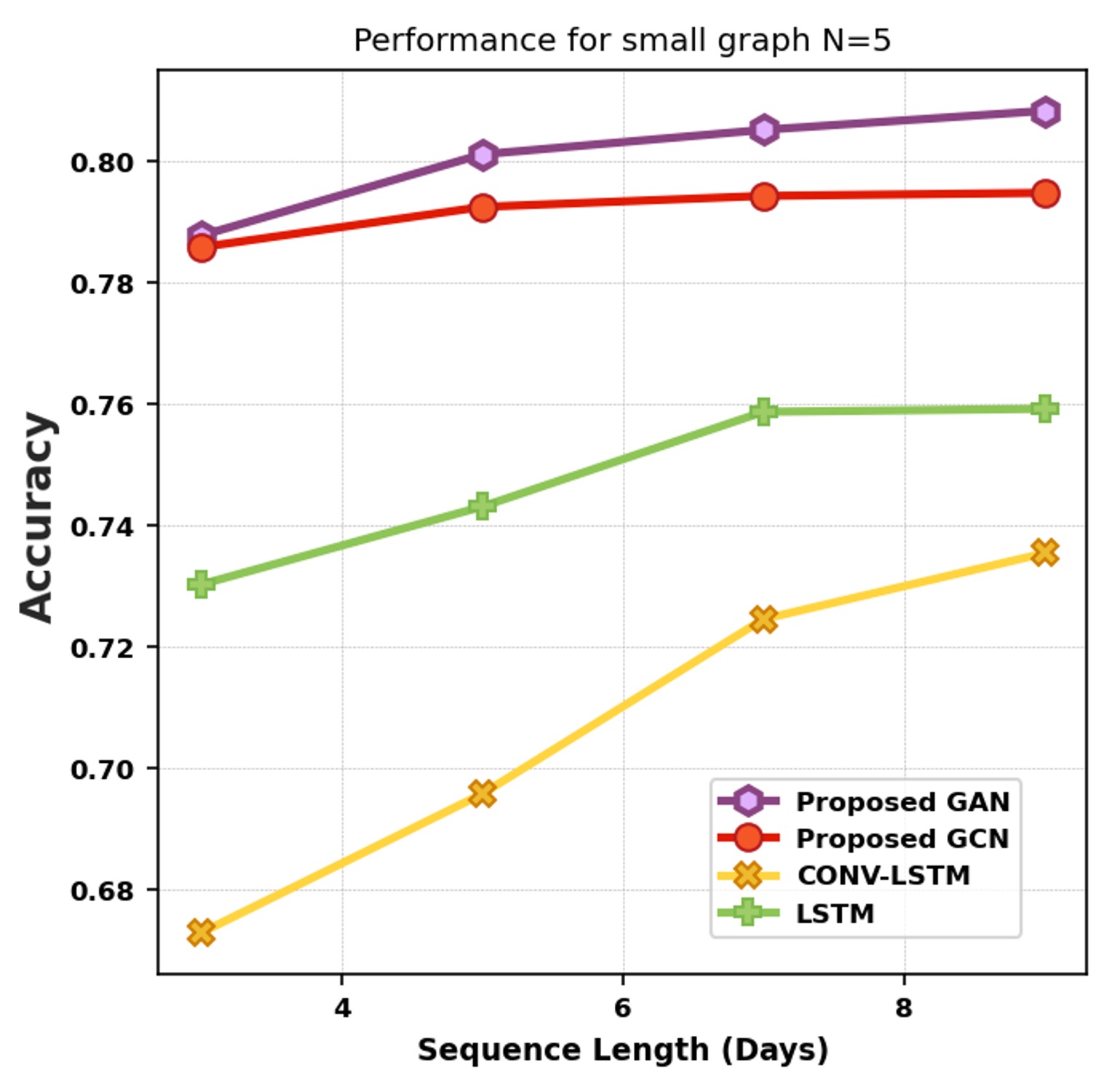} }}%
	\caption{Impact of network size and sequence length on performance}%
	\label{fig:graph}%
\end{figure}

Figure \ref{fig:graph}(a) shows the impact of network size with a short temporal memory of 3 days. Initially, increasing the network size improves performance for all 4 models. However, for GAN and GCN, the performance plateaus at N = 15 before dropping due to irrelevant aggregation from unimportant nodes in larger graphs. The difference between GCN and GAN suggests that attention helps mitigate this effect. LSTM does not show significant improvement with larger graphs due to the lack of aggregation. CONV-LSTM improves as graph size increases, plateauing at N = 15. For a longer sequence length of 7 days (Supplementary material Fig. \ref{fig:extra_net}(a)), smaller network sizes (N = 5) benefit from longer memory, and attention-based models outperform GCN by a wider margin.

To explore the impact of temporal memory L, we fix the graph size and vary L from 3 to 9 days. In Figure  \ref{fig:graph}(b), for L = 3, graph aggregation significantly improves proposed models compared to benchmark models. As sequence length increases, benchmark models perform better while GCN's improvement is minimal. GAN shows the highest gain in performance. The results for a larger network are presented in Supplementary material Fig. \ref{fig:extra_net}(b).
When the network size is large, the graph-based models perform really well at $L$ =3, but longer memory has a detrimental effect. In summary, very large networks and long memory lead to unstable behavior due to excessive aggregation from irrelevant nodes and past memories.

\subsection{Feature Importance} 
For better interpretability and to gain insight into how much individual modalities contribute to the proposed model GAN-LSTM's predictions, we compute feature importance. Since each unit in deep networks applies a non-linear operation to multiple inputs, visualization of weights provides little insight into which feature contributes the most. Additionally, our input has two dimensions: features and the number of nodes. Therefore, we utilize saliency maps \mbox{\cite{saliency}} to compute which parts of the input contribute the most to the model's predictions. To this end, first, the model is trained, and then for each test input, a saliency map is obtained by computing a normalized gradient for the input. The average map is obtained by taking the average across all input samples. Since we are interested in feature importance only, we collapse the map across the number-of-nodes dimension by taking the mean. For generalization, we repeat this process using the LOCO split described in section 
\mbox{ \ref{ml-pipe} }and report the mean and $95\%$ confidence interval across all trials.

The feature importance analysis reveals distinct standout features within each modality. However, given the vast number of features, it becomes impractical to visually represent the importance of every individual feature. Therefore, we opt to showcase the top ten features from each modality. This approach allows us to compare the significance of different modalities and gain insights into individual features that exert the most influence.

\begin{figure}
    \centering
    \includegraphics[width=10cm]{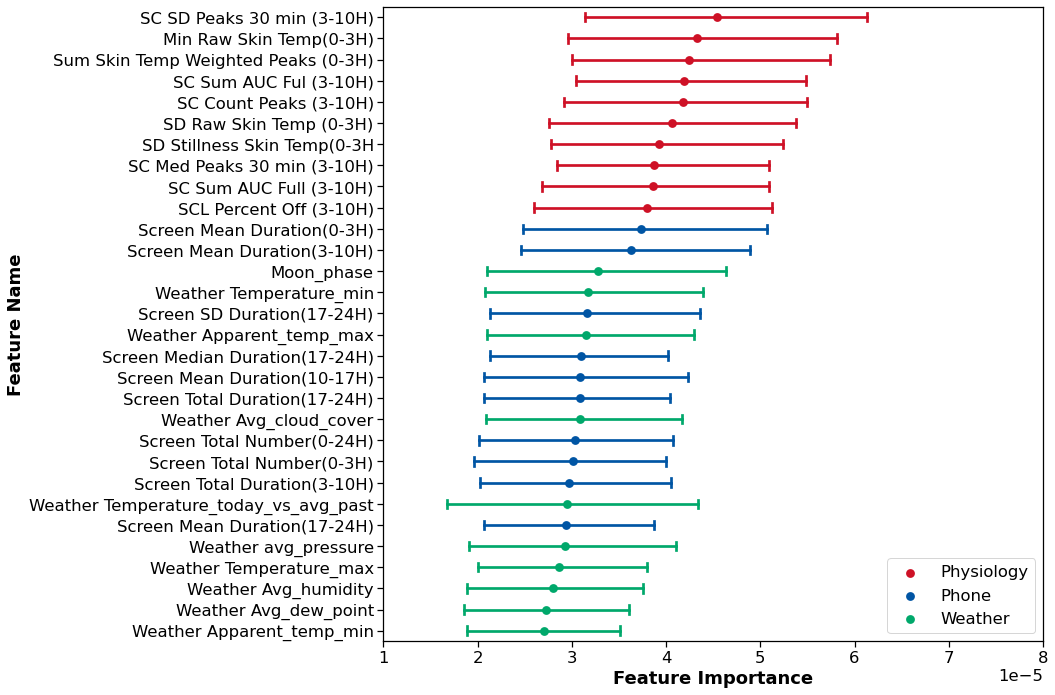}
    \caption{Feature Importance for different modalities for N=10 and L=3 days. \\ SC, SCL, and SD stand for Skin Conductance, Skin Conductance Level, and Standard Deviation respectively. \\ More details on features, in Tables \ref{tab:wearable-feat},\ref{tab:phone-feat}, and \ref{tab:weather_des}. }
    \label{fig:feat-imp}
\end{figure}

The top features from each modality are presented in Fig. \mbox{ \ref{fig:feat-imp}}. Comparing different modalities, physiological features hold the most importance. In particular, skin conductance (SC) and skin temperature during or just before sleep are the most significant.  Among phone features, phone screen usage patterns emerge as the most influential. Additionally, weather attributes like moon phase, cloud cover, temperature, and humidity exhibit similar levels of importance.

We further investigate the impact of the temporal module by removing it and observing the drop in performance. Our findings highlight that the temporal module aids in improving performance in proposed architectures and its impact is more pronounced in GCN-LSTM than GAN-LSTM. More details in Supplementary Material section \ref{temporal}.

\section{Robustness Analysis against Data Perturbations}
 Graph-based architectures exploiting contagion to complement the prediction model can expose the model to vulnerabilities caused by noisy or missing data. There are two key challenges in the development of a robust system in this context. First, wearable and mobile sensors provide high-resolution multi-modal data allowing us to train the system on a large feature space. However, in real-world scenarios, the larger feature space is more prone to sensor noise and missing data. Secondly, noisy graphs with irrelevant edges can also degrade the performance by aggregating information from nodes with missing or noisy data. To investigate the impact of perturbations in the data,  we conduct a systematic series of experiments.

To this end, we create artificial perturbations in the testing data from multiple dimensions and utilize different metrics to identify how different parts of the network are affected by these perturbations. For comparison, we repeat the same process for benchmark models defined in the previous section and report their performance.

\subsection{Artificial Perturbations}
\begin{figure}[t]
	\includegraphics[width=15cm]{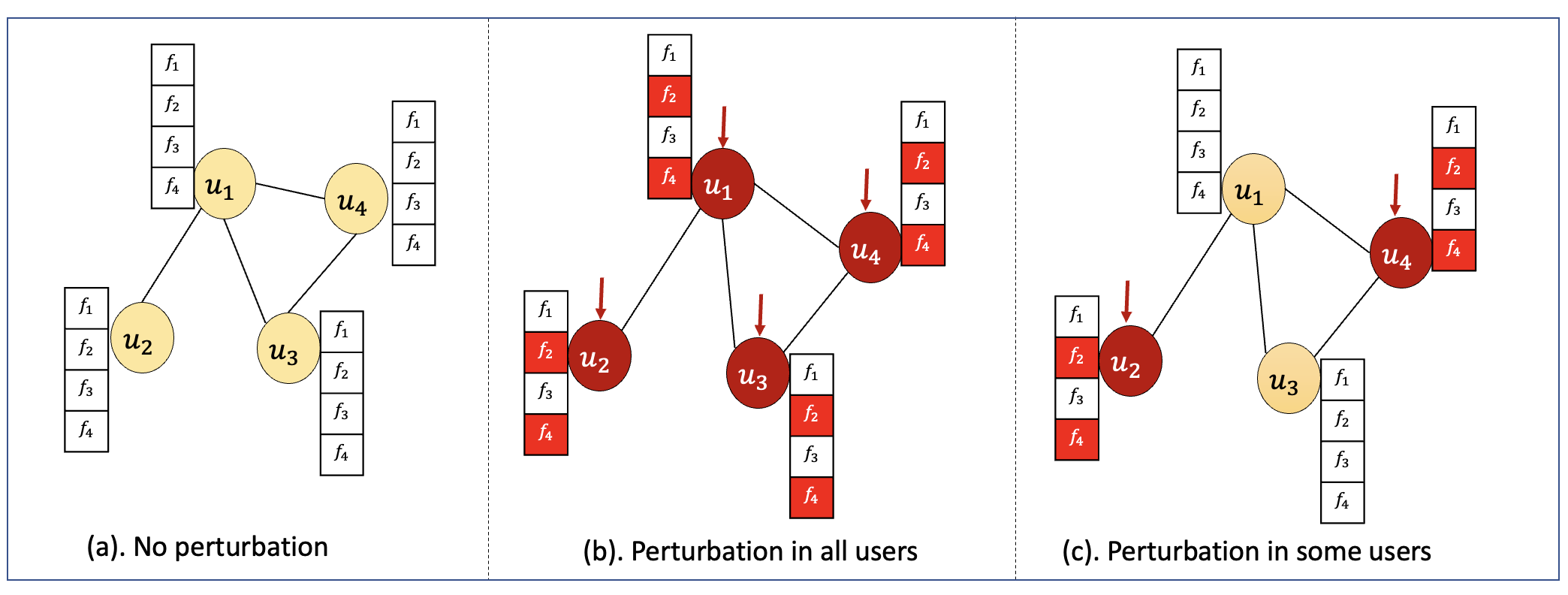}
	\caption{Different scenarios system perturbation. The circles represent nodes/users in the graph and the corresponding array represents the feature vector. The red arrows and red nodes represents perturbed participants. Yellow nodes have no perturbation. (a) The original system without any perturbation. (b) Randomly chosen 2 features $f_2$ and $f_4$ in red color are perturbed in all participants. (c) Randomly chosen features in 2 randomly chosen participants $u_2$ and $u_4$ are perturbed.}
	\label{pert}
\end{figure}
The data are perturbed at two levels, feature level and participant level. For feature level, we assume that the feature $f_i$, where $i\in \{1,2,..m\}$, follows an  univariate Gaussian distribution, $\mathcal{N}(\mu_{f_i} ,\sigma_{f_i}   )$. After estimating the distribution parameters from data, perturbations are created by replacing feature data with randomly sampled values from the learned distribution with extended standard deviation. Note that if we create aggressive perturbations by replacing data with values that the feature does not often take, then those would be detected as outliers and filtered out in the preprocessing phase. Instead, we create noise and/or missingness by imputing values that would not be detected in the preprocessing phase or would be actually used for imputation if data are missing.

We learn the parameters $(\mu_{f_i},\sigma_{f_i})$ of this distribution (mean and standard deviation), using the maximum likelihood estimation (MLE) \cite{prob}. The derivation for the estimated parameters is provided in supplementary material \ref{robust-noise},
%
%
We expand the distribution slightly by increasing the variance by a factor of three for generalization.
Once the distributions of features are learned, we investigate three cases. For ease of understanding, let us overview the original state of the system as shown in Fig. \ref{pert}(a). The yellow circles represent the participants $u_j$ connected in a network and each participant has an associated feature vector $x $ consisting of $f_i$ where $i \in \{1,2,..,m\}$.

 In the first scenario shown in Fig. \ref{pert}(b), we consider feature-level perturbation across all participants. First, a random subset $\tilde{I}_f$ of  size $\tilde{m}<m$ is selected from  feature index set $I_f = \{1,2,..,m\}$. Then, for each feature $f_p$, where $p\in \tilde{I}_f$, the values are imputed with randomly selected data from the corresponding distribution $\mathcal{N}(\mu_{f_p},\sigma_{f_p}   )$. This procedure is repeated for all participants in the network. In this example, the first 2 features are chosen at random, $f_2$ and $f_4$. Then the features are perturbed in all participants as shown by red participant nodes. 
 
 In the second case, we investigate a more complex scenario. Like previous scenario, first  a random subset $\tilde{I}_f$ of  size $\tilde{m}<m$ is selected from  feature index set $I_f = \{1,2,..,m\}$. Then a random subset of $\tilde{k}<N$ participants $\tilde{I}_u$ is chosen from the set of participants in the graph network $\mathcal{G}$ of size $N$.  In the end, perturbation is only applied to participants $u_q$ where $q\in \tilde{I}_u$. In both scenarios, perturbations are only applied to the current-day data, and past memory is not perturbed. In the third case, we vary perturbations along temporal memory.
  
\subsection{Quantifying Robustness \& Experiment Design}
To characterize the impact of perturbations on model performance, we utilize the accuracy score as the metric. It is important to randomize and repeat the process multiple times to counter sensitivity to the choice of features and participants. First, the model is trained on unperturbed training data. Then, the perturbation is applied to test data and the accuracy score is computed for all test samples. The data are randomly perturbed again $H_p$ times and in each iteration, the performance metrics are computed. 
The original trained model is sensitive to parameter initialization and training-test split of data. To alleviate this sensitivity, we repeat the process of random splitting, training a new model and testing it on perturbed data $H_t$ times.  Finally, for a given set of design parameters, the perturbation process is repeated $H_t \times H_p$ times. The average scores and $95\%$ confidence interval are reported.

To further investigate how network topology and a participant's location in the network impact its performance, we evaluate the importance of a node and its accuracy score. We characterize node importance by its centrality in the graph network \cite{networks}. To this end, we use two centrality metrics, degree, and eigenvalue centrality. Degree centrality assigns higher importance to nodes that have a large number of neighbors. However, it does not account for the cascade effect resulting from the fact that a node can also be important if it is connected to influential nodes. Eigenvalue centrality quantifies the influence of a node in a network by measuring the node's closeness to influential parts of a network. More details on these metrics are provided in Supplementary Material \ref{centr}.

The centrality metrics are categorized into three categories: alone, small, and large. Alone nodes are not connected to anyone or have very small eigenvalue centrality ($C_e<10^{-4}$) and are mainly affected by perturbations in their own features. Small centrality nodes have small degree ($0<C_d<4$) or eigenvalue centrality  ($10^{-4}<C_e<0.3$). Centrality scores greater than these thresholds are considered large. For each category, evaluation metrics are computed for a given trial and then average results over multiple ($\approx 500$) trials are reported. 

\section{Results for Robustness Analysis}
\subsection{Perturbations in a subset of features}

\begin{figure}[h]
	\centering
	\subfloat[][\centering Robustness against percentage of perturbed features in all participants. ]{{\includegraphics[trim=0 0 0 0,clip, width=0.45\textwidth]{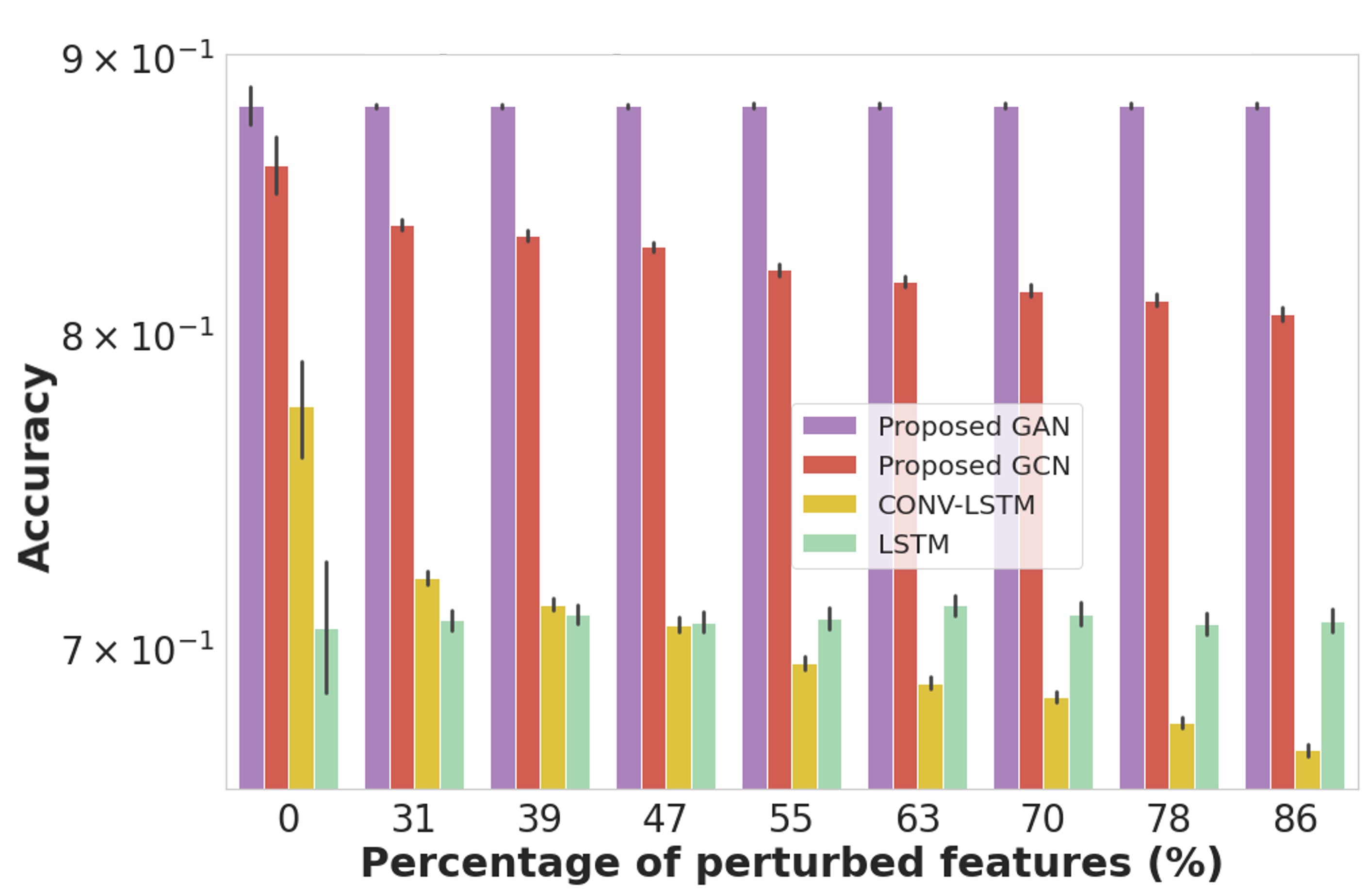} }}%
	\qquad
	\subfloat[][\centering  Impact on Accuracy caused by  perturbations in varying number of participants]{{\includegraphics[trim=0 0 0 0,clip,width=0.45\textwidth]{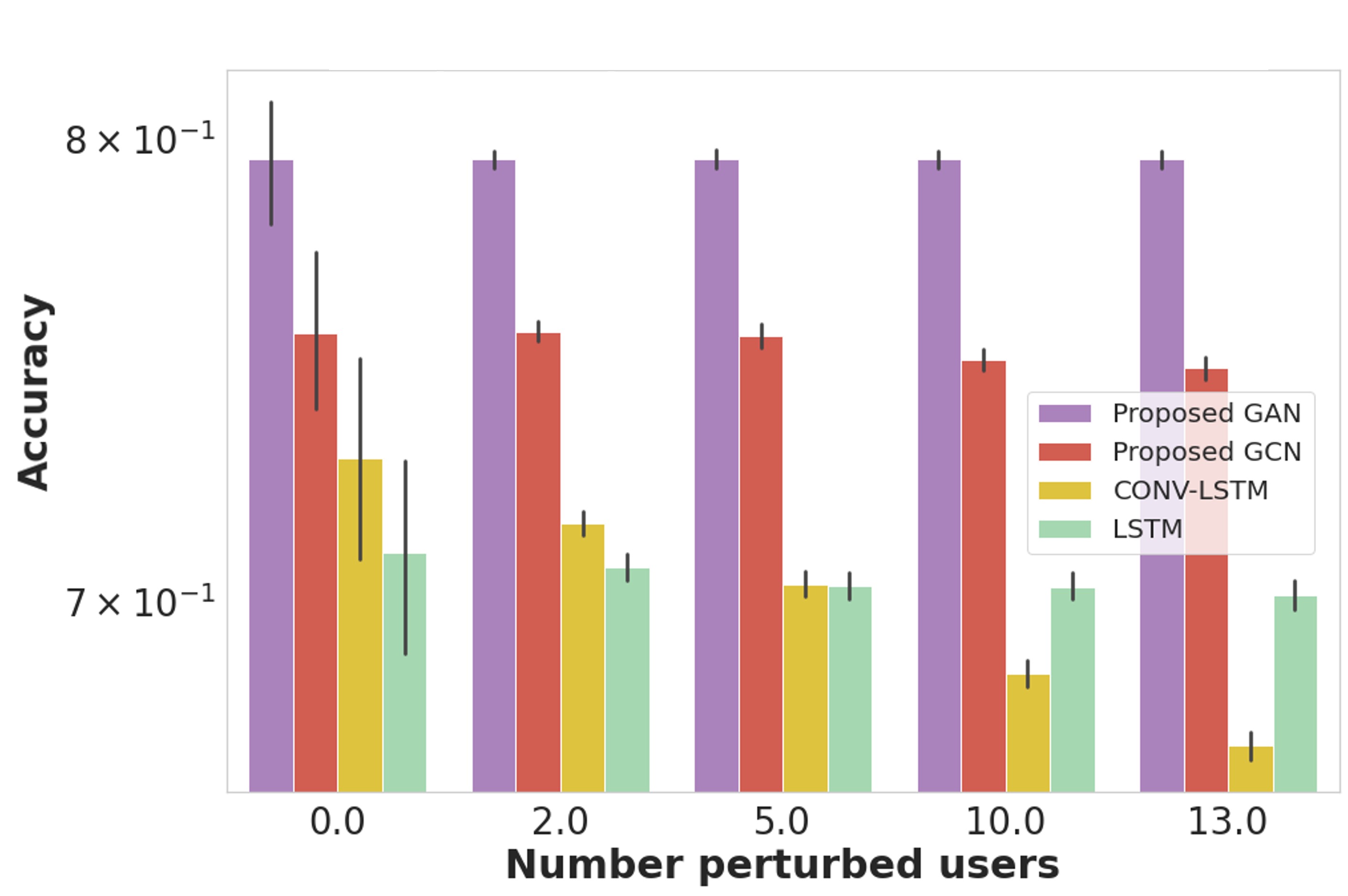} }}%
 \caption{ Impact of perturbations in a subset of features (a) and a subset of users(b).}%
	\label{fig:missing_feat}%
\end{figure}

In the first experiment, we perturb randomly selected features in inputs for all participants in a graph of size 15 and calculate performance. By increasing the percentage of perturbed features, we average the results over 500 trials, as shown in Fig. \ref{fig:missing_feat}(a). Features were perturbed in increments of 25 (e.g., 100, 125, 150, ...), and then converted to percentages for better understanding. At x=0 (no perturbation), all models perform well. As perturbations increase, performance drops, with LSTM showing overall poor performance. CONV-LSTM exhibits a sharp performance decline and is the least robust with numerous perturbed features. GCN also shows decreasing performance with more disturbance. However, attention-based GAN remains highly robust against perturbations.
\subsection{Robustness and Network Topology}
In the second experiment depicted in Fig. \ref{pert} (c), instead of all participants experiencing the same perturbation, a random subset of participants in a network is chosen.  
While aggregating information from multiple participants has shown promising improvement in prediction accuracy, it also makes the predictions of the entire network more vulnerable to perturbations if a few participants in the network have noisy/missing data.  Thus, we present how the proportion of perturbed participants impacts the performance in Fig. \ref{fig:missing_feat}(b) by perturbing $50\%$ of the features in a varying number of participants. As observed in Fig \ref{fig:missing_feat}(b) LSTM experiences very high errors consistently. CONV-LSTM shows a steep increase in error as more and more participants are perturbed.

\begin{figure}[h]
	\centering
	\subfloat[][\centering Robustness in different degree nodes when perturbation is applied to 5 participants]{{\includegraphics[trim=0 0 0 0,clip,scale = 0.195]{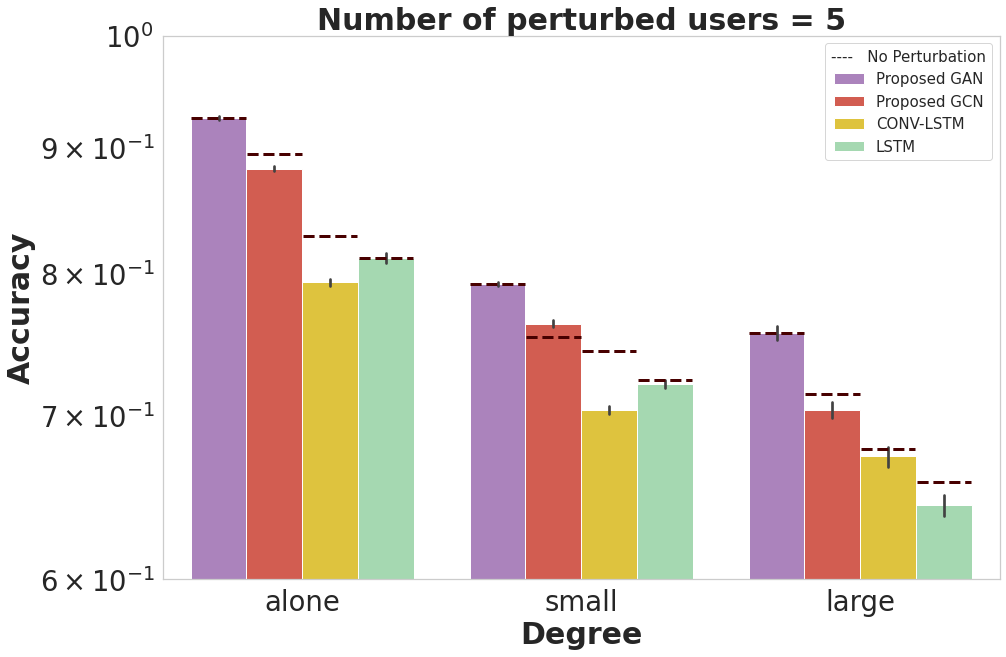} }}%
	\qquad
	\subfloat[][\centering Robustness in different degree nodes when perturbation is applied to 13 participants]{{\includegraphics[trim=0 0 0 0,clip,scale =0.195]{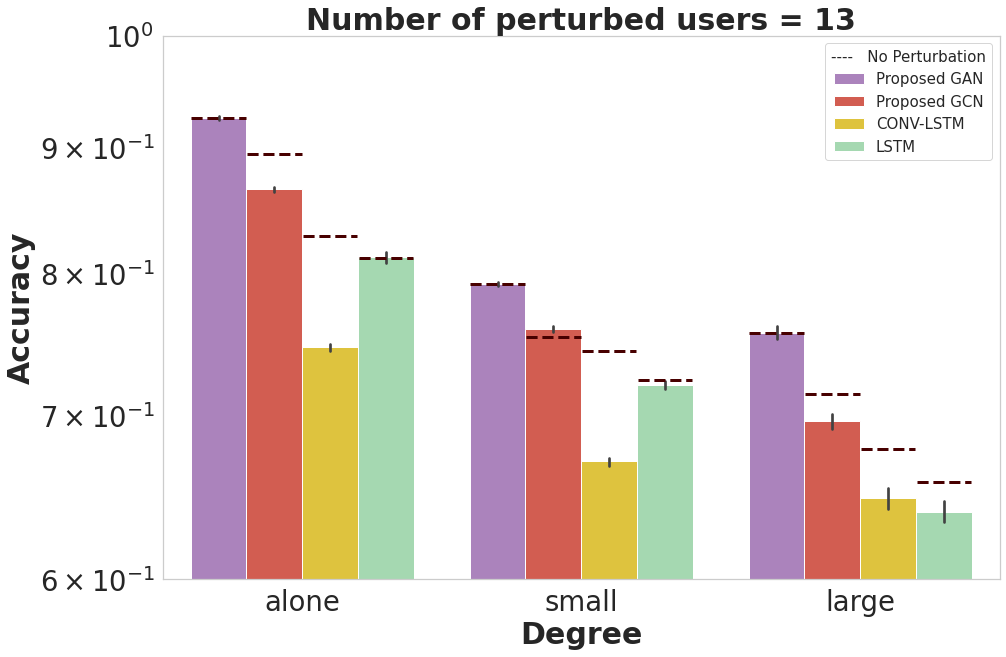} }}%
	\caption{Impact on the performance of different degree centrality nodes in the network by perturbing $50\%$ features in graphs of size 15 and sequence length of 3 days$.$ The dashed lines represent the average accuracy score without any perturbation$.$}%
	\label{fig:missing_deg}%
\end{figure}

The results presented in Figures \ref{fig:missing_feat} represent an average impact on the entire network. However, network topology also plays a role in how the model for some participants is more robust to noisy or missing feature data than others. In order to segregate these different parts of the network, we report the average accuracy score of different degree category nodes when 5 out of 15 participants experience perturbations in $50\%$  features, both chosen randomly, in Fig.\ref{fig:missing_deg}(a). The average performance without any noise for each model is represented by dashed lines. The category 'alone' indicates nodes that are not connected to anyone. The drop in performance in the 'alone' category results from the vulnerability of a participant to noise in its own features. Both GCN and CONV-LSTM suffer from a significant drop in performance for alone nodes. For nodes with a small degree, CONV-LSTM experiences a drop while other models are robust to it. Finally, for participants with a large number of direct neighbors, all models observe a drop in accuracy score because they have a higher chance of being closely connected to a node that is perturbed. We further increase the perturbations in Fig.\ref{fig:missing_deg} to 13  participants. The drop in performance is higher than what is observed in Fig.\ref{fig:missing_deg}(a), especially for large degree nodes.

\begin{figure}[h]
	\centering
	\subfloat[][\centering Robustness in different eigenvalue centrality nodes when perturbation is applied to 5 participants]{{\includegraphics[trim=0.5 0 3 0,clip,width=0.45\textwidth]{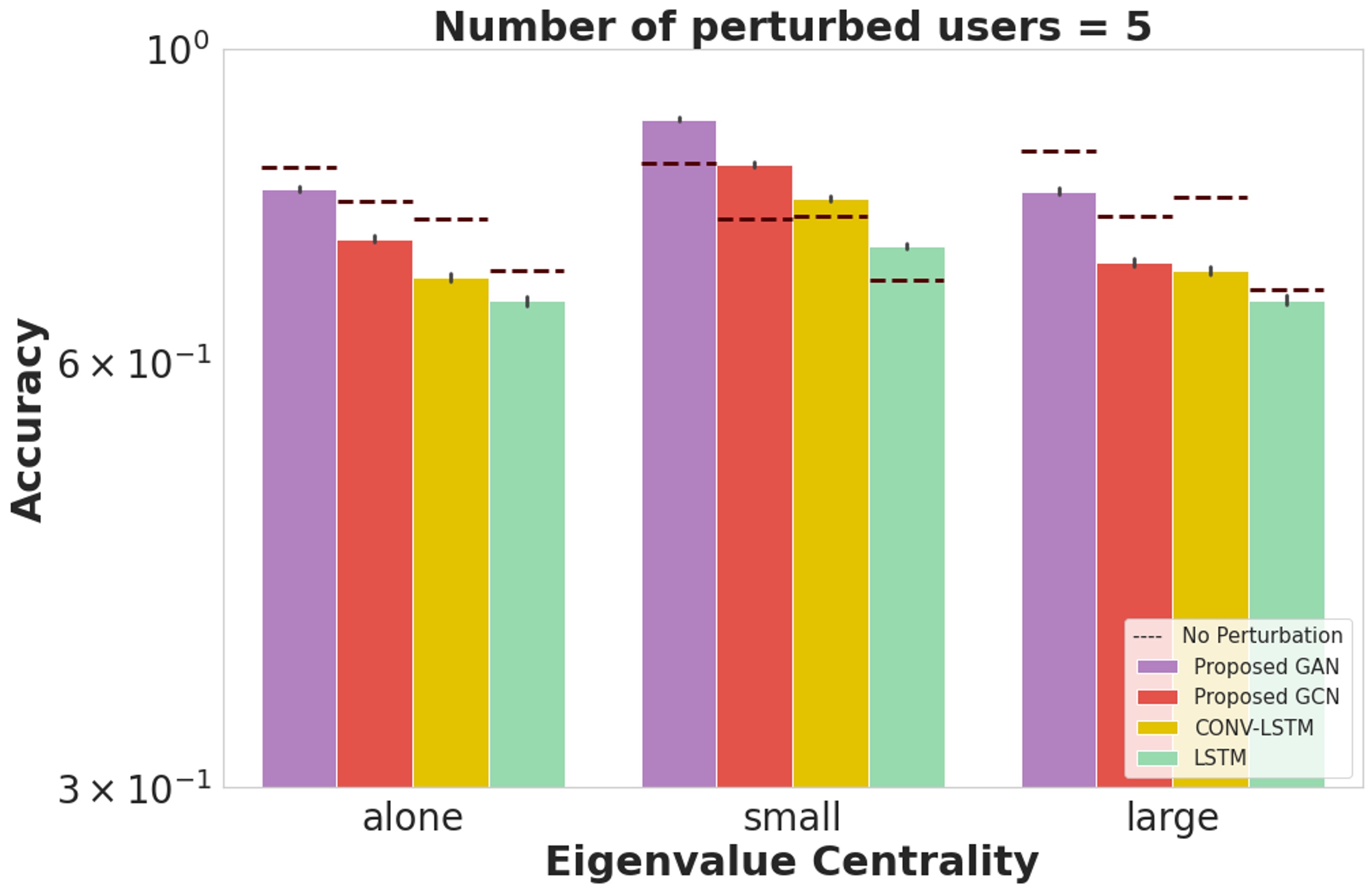} }}%
	\qquad
	\subfloat[][\centering Robustness in different eigenvalue centrality nodes when perturbation is applied to 13 participants]{{\includegraphics[trim=1 0 3 0,clip,,width=0.45\textwidth]{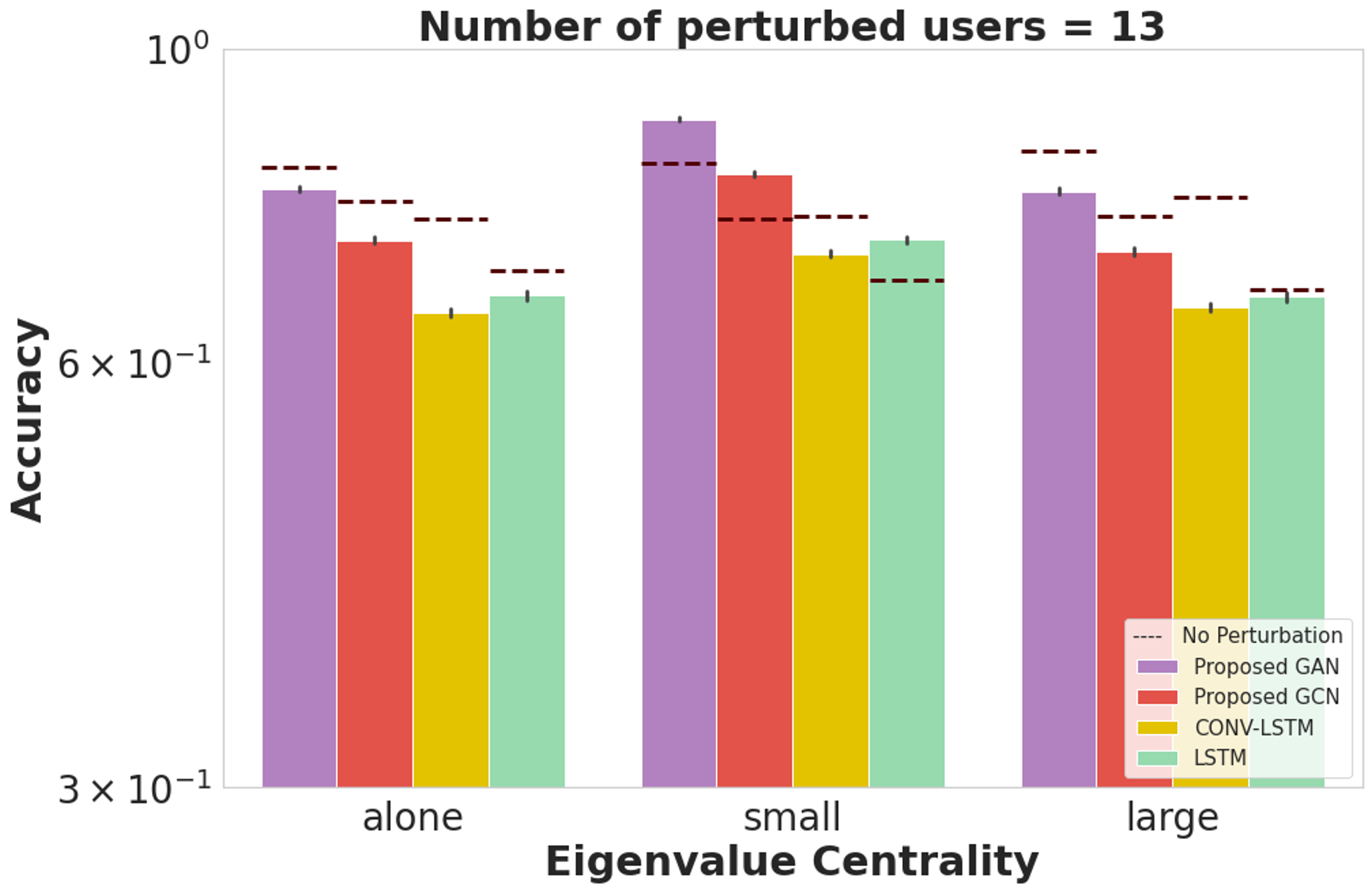} }}%
	\caption{Impact on the performance of different eigenvalue centrality nodes by perturbing $50\%$ features in graphs of size 15 and sequence length of 3 days$.$ The dashed lines represent the average accuracy score without any perturbation$.$}%
	\label{fig:missing_eig}%
\end{figure}

We use another centrality metric called eigenvalue centrality to segregate different parts of the network. First, a small number of participants i.e., 5 are perturbed in Fig.\ref{fig:missing_eig}(a). For participants who are alone or have extremely small eigenvalue centrality ($C_e<10^{-4}$), there is a sharp drop in performance. For small eigenvalues, only CONV-LSTM suffers from a drop in the accuracy score. For large eigenvalue, the behavior is unexpected and instead of a decrease, there is an increase in average performance. To investigate it further, we raised the number of perturbed participants to 13 (Fig.\ref{fig:missing_eig}(b)). For the 'alone' category, the drop in performance is slightly larger than before. For small eigenvalue centrality, we observe unexpected behavior, where performance is slightly improving for graph-based models. Finally, for large centrality nodes, there is a significant decrease in performance. This category is the only one in which even attention-based models are not robust. These results indicate that more connections to influential nodes can lead to unstable predictions.

Finally, we investigate how perturbations in different parts of the sequence impact the performance. Performance declines linearly for CONV-LSTM and GCN with increasing sequence perturbations, while attention-based models demonstrate robustness to these changes.Detailed results are provided in Supplementary section \ref{pertT}.

\section{Discussion}

\subsection{Unlocking the Potential of Contagion}
While existing literature has provided evidence for contagion through statistical testing \cite{NIHstudy}, there exists a notable gap in the field: a lack of quantitative modeling that hinders its seamless integration into learning systems. Our proposed models have the capability to discern and incorporate underlying contagious behaviors in the prediction task. This is especially valuable because of the inherent scarcity of health and wellbeing labels, such as user emotions, in contrast to the high-resolution features readily obtainable through wearables and phones.  This sparsity often necessitates some kind of downsampling of features, resulting in information loss. Because of the same limitation in our dataset, we were constrained to extracting statistical descriptions from multiple features rather than utilizing raw data. Leveraging contagion through graph networks offers a novel approach to harnessing the potential of wearable and smartphone data. This approach provides complementary information to the model, effectively reducing the loss of valuable information.
\subsection{Enhancing Sleep Duration Prediction through Data Aggregation}

While comparing different models' performance in Table \ref{tab:eval}, we observe an improvement when the model aggregates data from multiple participants in a certain pattern, e.g. when employing CNN. However, the extent of this gain is limited due to the absence of information about the connections between these participants. Thus, the inclusion of graph networks significantly enhances the model's predictive capabilities, providing indirect evidence of contagion. Nonetheless, it is crucial to acknowledge that not all friends or neighbors contribute equally, nor do they exert an equal magnitude of contagion. Therefore, when the model focuses its attention solely on important nodes using GAN, we not only witness performance improvements but also a substantial increase in model robustness as shown in  Fig. \ref{fig:missing_feat} and \ref{fig:missing_eig}.
 
 Our framework paves the way for testing different networks in digital health systems. Graphs provide a simple yet powerful representation of multiple participants and links between them that can be exploited in numerous ways.  In addition to the social network, graphs can be constructed from similar lifestyle patterns\cite{mob} (e.g., mobility, eating, and exercise habits)  to complement the multi-modal prediction. The architecture proposed in this work can handle temporal graphs as well as those with dynamic sizes. This means that as the user's social interactions change, the model is able to adapt to the changing contagious behavior. 
\subsection{Robustness and Node Influence}
Robustness analysis highlighted that graph integration helps make the model more robust to data perturbations only when attention is paid to important nodes. Relying solely on the graph neural network for combining information from other nodes can make some nodes more vulnerable to noisy or missing data. Interestingly, the impact of perturbations on the performance varies depending on a node's connectivity. As shown in Fig. \ref{fig:missing_deg} and \ref{fig:missing_eig}, nodes directly connected to a substantial number of peers experience a less pronounced decline in performance compared to those linked to a select few highly influential nodes.  Firstly, nodes with higher degrees exhibit lower accuracy relative to their counterparts with fewer connections. The attention mechanism helps mitigate this effect to some extent and also makes the model more robust to perturbations. However,  the users with high eigenvalue centrality experience a large drop in performance even with GAN. Thus, not all neighbors are equal, and different friends impact a node's performance differently based on their own influence in the network.

\subsection{Optimizing Network Size }
Network size plays a pivotal role for two reasons: firstly, we want to capture all nodes that might have a contagious effect on user's behavior. Therefore, we vary the graph size from 5 to 20 and observe the performance in Fig. \ref{fig:graph}(a). Our findings revealed a performance plateau at N=15. Larger graphs tended to introduce irrelevant data aggregation, although this could be partially alleviated by the implementation of an attention mechanism. This tradeoff can be optimized by carefully designing network size which is customized based on the user's sociality, demographics, and lifestyle. For instance, in our study, all participants were university students who knew other people in the study; for these participants, we expect interactions with a larger number of fellow students. However, for specialized groups like caregivers, who may have extensive interactions within smaller circles (mainly care receivers), a smaller network size may be required for reliable performance. Graph-based models tend to be more sensitive to perturbations in neighboring nodes' features due to their aggregation mechanism. 

This challenge can be addressed using GEDD which enables us to optimize network sizes, thereby mitigating the adverse effects of large networks and noisy data while generating graphs tailored to the available dataset.

\subsection{Harnessing the Power of Attention}

While constructing social networks is relatively straightforward, their effective use in health label prediction presents multiple challenges. This is exemplified in Figs.  \ref{fig:missing_deg} and \ref{fig:missing_deg_t}, where GCN performs poorly for nodes with a large number of direct neighbors that are also more susceptible to perturbations. However, the introduction of attention mechanisms significantly mitigates irrelevant data aggregation, resulting in improved performance. This underscores the critical role of attention in our approach. The attention mechanism in this work is based on feature correlation. It can be further improved by utilizing complementary information from study surveys or customized based on network topologies.  
We noticed in our analysis that nodes with large eigenvalue centrality are more prone to noise. This insight can be used to do a weighted normalization of attention weight or thresholding the maximum number of important nodes when the node degree is very large.

\subsection{Limitations}
The collected dataset was a human-centric dataset posing challenges such as a limited number of participants. This limitation makes it impossible to capture the entire social network of each participant  (e.g., friends, family, coworkers) in one study. Therefore, the maximum graph size in our evaluation is only 20, which does not represent very complex graph typologies. However, despite these limitations, the proposed architecture shows promising results and highlights the potential of social network integration in digital health systems. While this work focuses only on sleep duration, it can be easily extended to other health and well-being labels such as mood, emotion, and eating habits. Another limitation of this dataset is that all the participants were young adults at one university which results in sampling bias. In future works, the system needs to be tested with other diverse groups of the population. Moreover, the acquisition of ground truth involved user questionnaires, a method susceptible to subjective biases including recall limitations. These inherent biases can be mitigated in future works through the integration of sleep-tracking technologies \cite{dopple_sleep}, such as smartwatches and under-mattress sleep trackers.
 The study was limited to metadata about calls, SMS, and screen on/off only. In future works, data from messenger  or social media apps can be utilized to further expand the social networks used in the model.  

Another limitation of this system is that it is completely centralized. Information from all users is directed to one centralized model for making predictions. In order to deploy this system in large-scale real-life networks, this issue should be resolved through an elaborate distributive computing solution\cite{decentral}. Furthermore, the robustness analysis makes assumptions about the distribution of features in order to perturb them. We also used similar assumptions for standardization and removing outliers in data preprocessing. In future works, this assumption can be relaxed and domain knowledge about feature distribution can be utilized to further improve this analysis.

\subsection{Future Work}

\begin{itemize}
    \item  \textit{Selective Sampling: } In multi-user architecture, complete feature information about all users is required by the system. As observed in the robustness analysis, the proposed model can produce reasonable performance even when some users are missing over $50\%$ data. The drop in performance strongly depends on where in the network a node is located. This insight can be extended in future work to identify \textit{influential} nodes in the network. These nodes can either be tapped for the collection of existing modalities or additional information that can help improve the performance of the entire network.

\item \textit{Active Mobile Health: }In systems where health labels are monitored, tracked, or predicted from non-obtrusive data sources, the role of the system is passive. On the other hand, in an active mHealth system, in addition to monitoring and predicting, the system can also provide interventions to regulate the behavior of interest. Scaling this to a network level, where the users are interconnected and exert influence on each other's behaviors, important nodes can be identified to provide possible points of intervention for regulating the behavior of the entire network.

\item \textit{Lifestyle recommendations: } The proposed system can be extended to provide lifestyle recommendations to rectify unhealthy behaviors. A unique feature of this extension would be that in addition to changes in daily habits, it can also give recommendations about social interactions conducive to enhanced health and overall wellbeing.  Such recommendations would need to be tested before implementation.
\end{itemize}

\section{Conclusion}

In this work, social networks are integrated with ubiquitous mobile and wearable sensor data to predict next-day sleep duration labels ( i.e., more or less than 8 hours) for multi-user networks. To achieve this goal, an architecture based on graph convolution networks is developed that aggregates information from different parts of the network. The proposed architecture showed distinct gains in prediction performance, corroborating the existence of contagion in sleep behavior and highlighting the potential of social networks to complement pervasive sensing data in mobile health systems. Further, robustness analysis exhibited the stability of the proposed architecture to perturbations in data. It also highlighted the significance of network topology as nodes with lower eigenvalue centrality are less impacted by disturbances compared to those with higher connectivity to influential nodes. This work provides a practical and scalable approach, to identifying dynamic social networks and integrating them into prediction models for health labels. In addition to performance evaluation, a systematic analysis to characterize the system performance when deployed in the wild is also conducted. The robustness analysis exhibited the stability of the proposed architecture to perturbations in data. It also highlighted the significance of network topology as nodes with lower eigenvalue centrality are less impacted by disturbances compared to those with higher connectivity with influential nodes.

\section*{Acknowledgments}
This work was supported by NSF (\# 1840167 and \# 2047296), NIH (R01GM10518), Samsung Electronics, and NEC Corporation.
We thank SNAPSHOT study participants and collaborators.

\bibliographystyle{ACM-Reference-Format}
\bibliography{base1}
\appendix

\section*{Supplementary Material}

\section{Data Preprocessing}\label{preprocess}
The data are processed in three steps,
\begin{itemize}
	\item \textbf{Processing missing data: }
 Data collected from various sources over an extended period often contain missing values, particularly in wearable and mobile data due to factors like battery drainage, app issues, and participants not wearing the wearable. We addressed this in three steps. First, we discarded features with more than half of their values missing. Then, we separated data for each user and used their available data to fill in missing values, employing k-nearest neighbor (kNN) imputation \cite{imputation}. KNN identifies the $\tilde{k}$ nearest neighbors based on Euclidean distance. Finally, we combined data from all users and performed KNN imputation across users to complete the data.

	\item \textbf{Removing outliers: } After processing missing information, outliers resulting from sensor noise and mobile app glitches are removed. We detect outliers using Z-score statistics with a cutoff of four. A Z-score is computed by subtracting the overall mean from the datapoint and dividing by the overall standard deviation.
	
	\item \textbf{Standardization: } 
 The dataset contains multiple modalities with different distributions, so we standardize the data to have a zero mean and unit standard deviation \cite{z-score}. This standardization aids gradient-descent optimization by smoothing the landscape \cite{z-scoreGD}. In this process, similar to the Z-score, we subtract the mean from the data and divide it by the standard deviation. Importantly, we only learn standardization parameters from the training data to ensure fairness. We randomly split the entire dataset into training and test data, learn parameters from the training data, standardize the training data, and then apply the same parameters to normalize the test data without looking at the test data to maintain fairness in the machine learning pipeline.

\end{itemize}

\section{Robustness analysis additional information}

\subsection{Centrality Metrics}\label{centr}
\begin{itemize}
\item \textbf{Degree Centrality} is a direct representation of how many directly connected neighbors a node has. If the graph is represented by $N\times N$ adjacency matrix $A$, then the degree centrality $C_d$ of  a node $v$ is calculated as,
\begin{equation}
	C_d(v) = \sum_{u=1}^{N}\mathcal{I}(A_{uv})
\end{equation}
where, 
\begin{equation}
	I(x) = \begin{cases}
		1 & \text{if  $x > 0$}\\
		0 & \text{x = 0}
	\end{cases}   
\end{equation}

Degree centrality assigns higher importance to nodes that have a large number of neighbors. However, it does not account for the cascade effect resulting from small degree nodes connected to a few but influential nodes.

\item \textbf{ Eigenvalue centrality} quantifies the influence of a node in a network by measuring the node's closeness to influential parts of a network. It combines the degree of a node with the degree of its neighbors. For a graph $\mathcal{G}$ with adjacency matrix $A$, the eigenvalue centrality $C_e$ of a node $v$ is calculated by \cite{networks},
\begin{equation}\label{eigen}
	C_e (v)=  \alpha \sum_{u,v\in \mathcal{E}}C_e(u) \;\;\; , \;\;\; AC_e = \alpha^{-1}C_e
\end{equation}
where $C_e$ and $1/\alpha$ are the eigenvector and corresponding eigenvalue of $A$ respectively.

\end{itemize}

\subsection{Noise Distribution}\label{robust-noise}

We learn the parameters $(\mu_{f_i},\sigma_{f_i})$ of the feature distribution (mean and standard deviation), using the maximum likelihood estimation (MLE) \cite{prob}. 
The likelihood function for MLE maximizes the likelihood of observing feature data $X_{f_i}$ given the distribution parameters $(\mu_{f_i},\sigma_{f_i})$, 
\begin{equation}
\mathcal{L} = P(X_{f_i} |\mu_{f_i},\sigma_{f_i} ) \;\;\; , \;\;\; \mathcal{L} = \mathcal{N}(X_{f_i} |\mu_{f_i},\sigma_{f_i} )
\end{equation}
We estimate the parameters by maximizing the log of the likelihood function. Since we assume the data samples are independent, the log allows us to take a summation over the  samples indexed by $v$ and represented by $x_i^v$,

\begin{equation}
	\hat{\mu}_{f_i} = \underset{\mu_{f_i}}{\mathrm{argmax}}\,  \sum_{v} log \big( \mathcal{N}(x_i^v |\mu_{f_i},\sigma_{f_i} ) \big)  \;\;\; , \;\;\; \hat{\sigma}_{f_i} = \underset{\sigma_{f_i}}{\mathrm{argmax}}\,  \sum_{v} log \big( \mathcal{N}(x_i^v |\mu_{f_i},\sigma_{f_i} )\big)
\end{equation}
Plugging in the expression for Gaussian distribution, expanding the terms as summation followed by the second derivative equated to zero, provides the following expressions,
\begin{equation}
		\hat{\mu}_{f_i} = \frac{1}{V} \sum_{v} x_i^v  \;\;\; , \;\;\; \hat{\sigma}_{f_i} = \frac{1}{V} \sum_{v} (x_i^v - \hat{\mu}_{f_i})
\end{equation}


where $V$ represents the total number of samples. 
Finally, we expand the distribution slightly by increasing the variance by a factor of three for generalization.

\section{Additional Results}\label{add}

\subsection{Impact of Network Characteristics}

\begin{figure}
	\centering
	\subfloat[][\centering Impact of graph size with long sequence]{{\includegraphics[width=5.5cm]{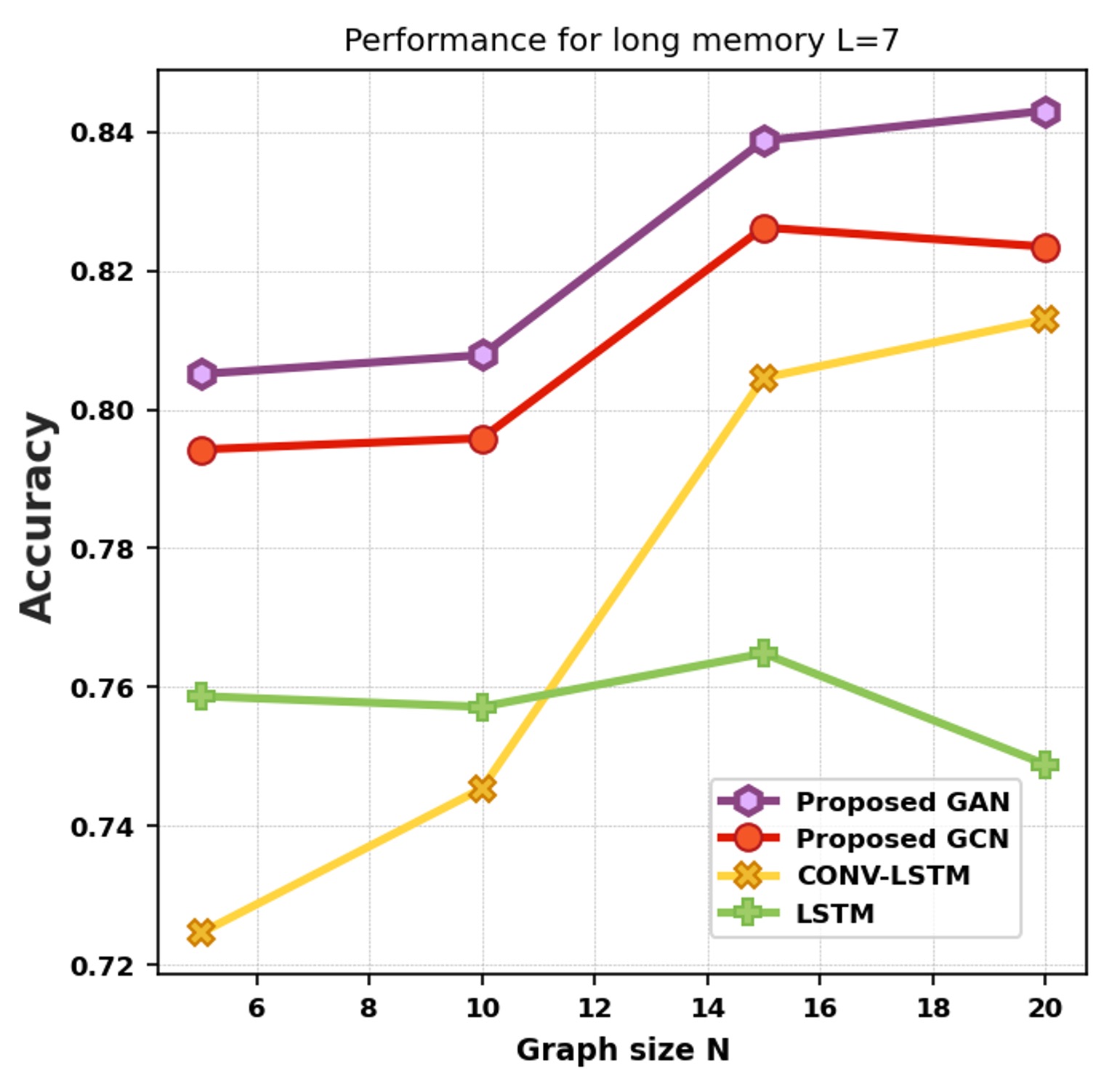} }}%
	\qquad
	\subfloat[][\centering Impact of temporal memory with large network]{{\includegraphics[width=5.5cm]{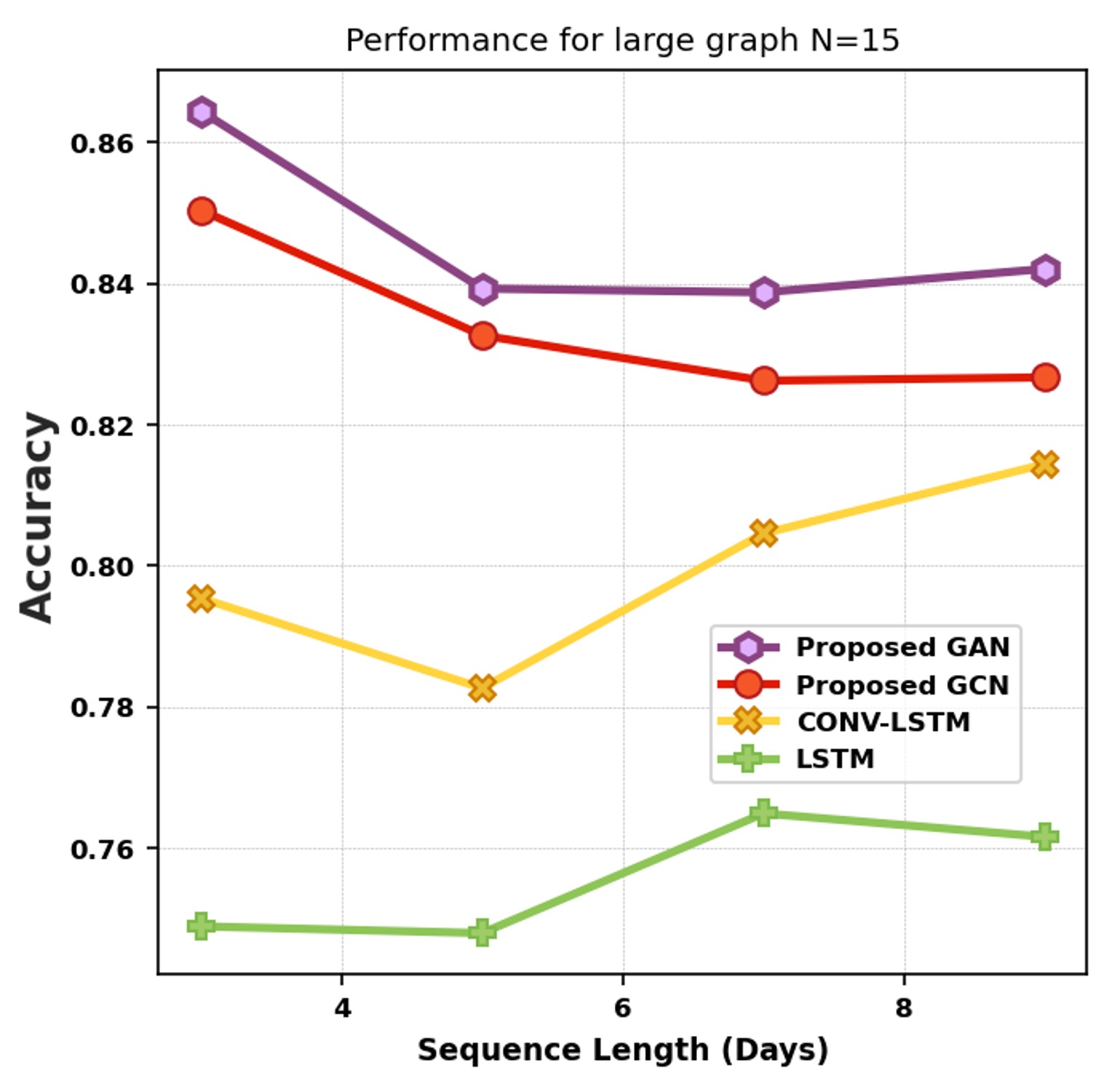} }}%
	\caption{Impact of different parameters on  performance}%
	\label{fig:extra_net}%
\end{figure}

We conducted another experiment to assess the impact of network size on the model's performance with an extended sequence length of $L=7$ days, and the results are shown in Fig. \ref{fig:extra_net}(a). Similar to previous findings, increasing the network size enhances performance until it plateaus at approximately $N=15$  for the proposed models. GCN experiences an accuracy dip after $N=15$, while GAN and CONV-LSTM continue to improve. We also repeated the experiment with varying sequence lengths while keeping the network size fixed at 15, as depicted in Fig. \ref{fig:extra_net}(b). Graph-based models perform exceptionally well at $L=3$, but longer memory adversely affects their performance. Conversely, benchmark models initially dip or show a marginal change in performance, but they improve significantly when memory extends beyond 5 days.

\subsection{Robustness Analysis}

\subsubsection{Perturbations in temporal domain}\label{pertT}

In this section, we present results for perturbations in a different part of the sequence. Please note that in all previous experiments, the sequence index was kept fixed and perturbations were only added to data from the current day while the prediction was being made for the next day. The feature matrix $S_k^L$, fed to LSTM module consists of past $L$ days of data,

\begin{equation}
	S_k^L =\big[x[k-L]] ,x[k-L+1],...x[k]\big]
\end{equation}

\begin{figure}%
	\centering
	\subfloat[][\centering Impact on error caused by  perturbations in varying number of days]{{\includegraphics[width=0.47\textwidth]{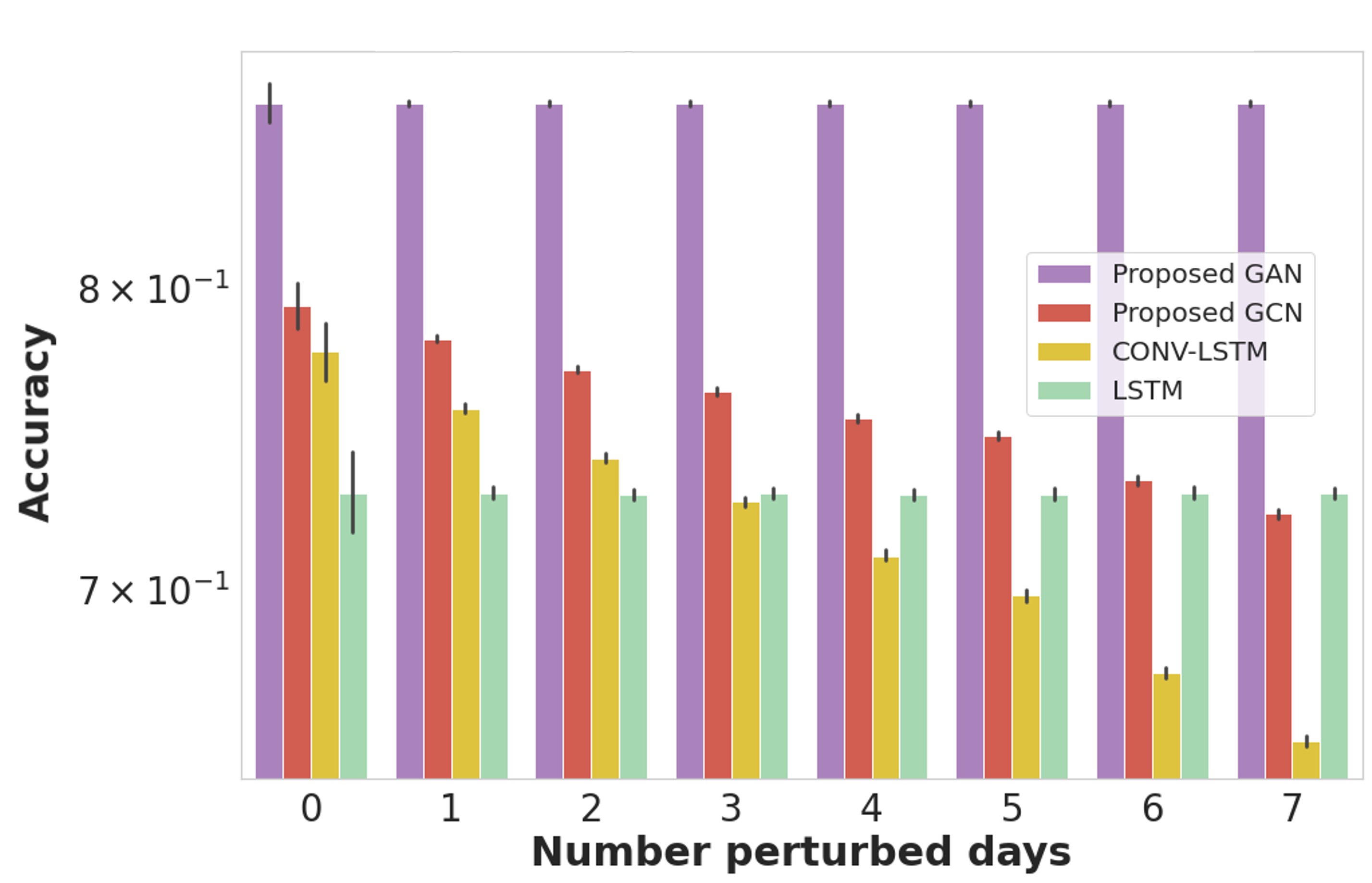} }}%
	\qquad
	\subfloat[][\centering  Impact on Mean absolute percentage error (MAPE) caused by  perturbations in varying number of days]{{\includegraphics[width=0.47\textwidth]{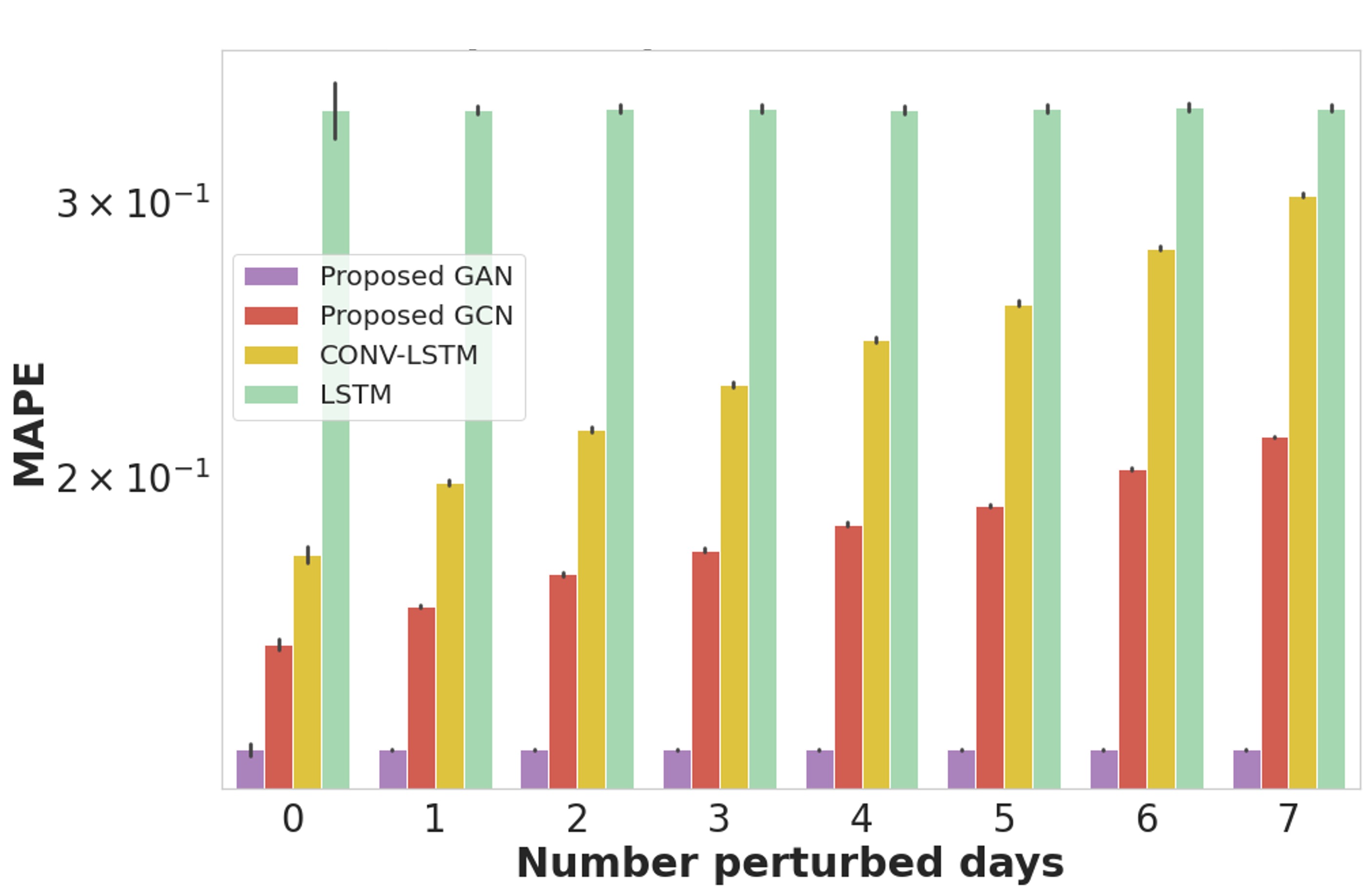} }}%
	\caption{ Impact of perturbation in  $50\%$ features in varying sequence length for all users }%
	\label{fig:missing_seq}%
\end{figure}

We randomly choose a subset of indices $\tilde{I}_L$ of size $\tilde{L}<L$ from the set $I_L={1,2,...L}$ and in those chosen days of sequence perturb a randomly chosen set of features. In Fig. \ref{fig:missing_seq}, we present results for a $50\%$ feature perturbation in a network of size 15 with a 7-day sequence length. Fig. \ref{fig:missing_seq}(a) displays the accuracy score, where LSTM maintains a consistently low score. CONV-LSTM experiences a sharp accuracy drop as more past data is perturbed, while GCN's performance linearly declines with an increasing proportion of perturbed days. The attention-based model proves robust to these perturbations.

\begin{figure}%
	\centering
	\subfloat[][\centering Robustness in different centrality nodes when perturbation is applied to 1 randomly chosen day for all users]{{\includegraphics[width=0.47\textwidth]{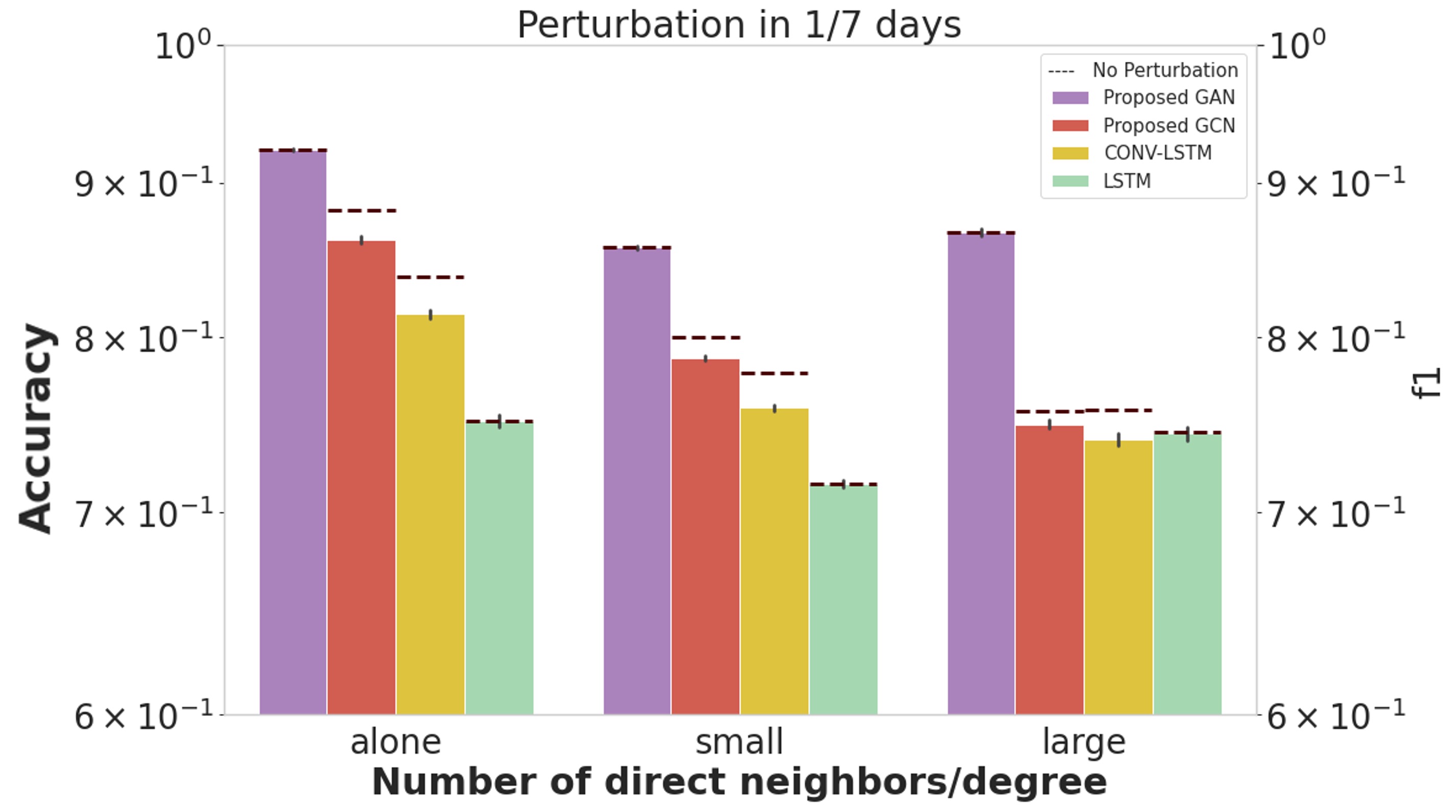} }}%
	\qquad
	\subfloat[][\centering Robustness in different centrality nodes when perturbation is applied to 5 randomly chosen day for all users]{{\includegraphics[width=0.47\textwidth]{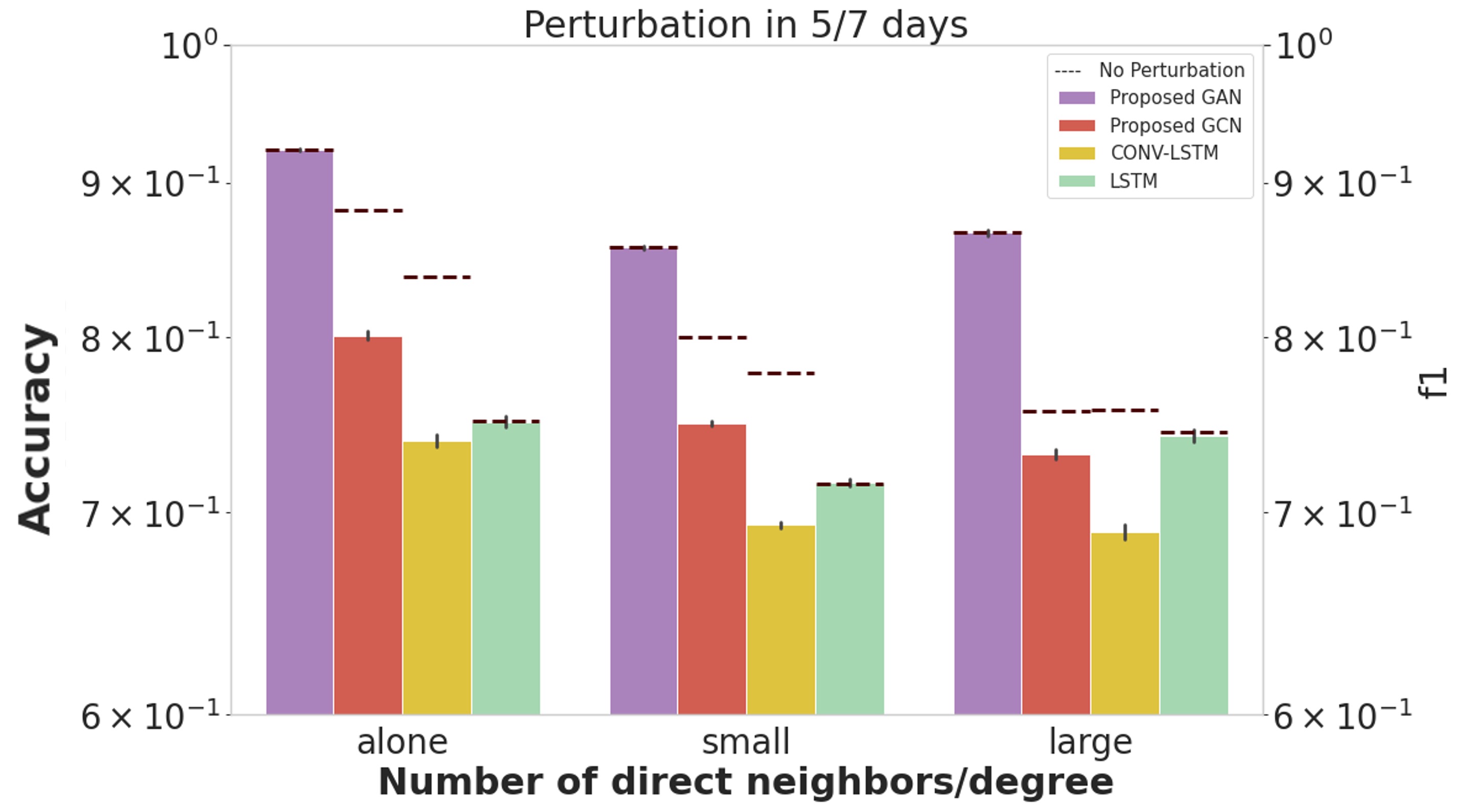} }}%
	\caption{Impact on the performance of different nodes in the network by perturbing $50\%$ features in graphs of size 15 and sequence length of 7 days$.$ The dashed lines represent the average accuracy score without any perturbation$.$}%
	\label{fig:missing_deg_t}%
\end{figure}

Fig. \ref{fig:missing_seq}(b) shows the absolute percentage error, with CONV-LSTM and GCN both exhibiting a linear increase in error as more days are perturbed, while GAN's performance sees a negligible drop. For a detailed examination of how different nodes are affected by varying proportions of perturbed days, refer to Fig. \ref{fig:missing_deg_t}.

\section{Impact of temporal module}\label{temporal}
The proposed models consist of two components: graph-aggregation (GCN/GAN) and temporal module. To depict the significance of the temporal module, we remove LSTM layers from the model and observe the change in performance for both proposed models, GCN-LSTM and GAN-LSTM, in the LOCO setting. First GCN-LSTM is trained and performance is computed on the test set. Then temporal module is removed and GCN-only model is trained on the same training set and performance is computed on the same test set.  Finally, the difference in performance as a percentage of the original model's (GCN-LSTM) accuracy is computed. The same procedure is repeated for GAN-LSTM and GAN only. This shows the contribution of the temporal module.  We conduct this analysis for every test-train split, calculating the percentage change in accuracy scores resulting from the omission of LSTM layers for the held-out cohort test set. 

For GCN-LSTM, on average the model's performance dropped by \mbox{$9\%$} with a standard deviation of \mbox{$3\%$}. For GAN-LSTM, the impact is much less significant with an average performance drop of around \mbox{$2\%$} and a standard deviation of \mbox{$1\%$}. There are two main findings of this experiment. Firstly, the temporal module contributes to the performance of both architectures. Secondly, the contribution of this module is more significant for GCN-LSTM than for GAN-LSTM. These findings align with the previously reported results where GAN-LSTM provided higher performance gains and exhibited greater stability than GCN-LSTM.

\section{Dataset} \label{supp}
\subsection{Dataset Description}\label{supp-data}

After preprocessing the data (details in Supplementary Material section \ref{preprocess}),   there were 3536 data samples available from 243 participants. 
There were available data for 15-25 days from most participants in cohorts 1-6 and for 40-60 days from most participants in the last cohort. This dataset is available at \cite{datasetSpand11}.

\subsubsection{Feature Description}\label{feat}
There are four types of data sources: wearable,  mobile phone, questionnaires, and weather. A detailed description of processed features obtained from these four sources is provided below.

\begin{itemize}
    \item Wearable device: The device worn by the participants on their wrist collected Skin Conductance (SC), Skin Temperature (ST), and 3-axis Accelerometer at 8 GHz. The units of Electro-Dermal Activity (EDA), ST, and Accelerometer were microsiemens ($\mu S$), Celsius ($^\circ C$), and gravity ($G$) respectively. EDA is a physiological measurement that assesses the electrical conductance of the skin in response to sympathetic nervous system activity\cite{EDA}. EDA consists of  Skin Conductance Level(SCL) and Skin Conductance(SC) Peaks (skin conductance response). SCL is the tonic electrical conductivity of skin while the peaks are phasic, short-term fluctuations in EDA due to a stimulus. In response to a stimulus, the EDA rises rapidly and then decays exponentially. These peaks are extracted using the procedure defined in \cite{EDA_Explorer}.  The data for one 24-hour day were divided into three 7-hour and one 3-hour windows from  3-10H, 10-17H, 17-24H, and 0-3H. These time windows were chosen based on participants' sleep routines. The average bedtime and waketime were 2:44 am ($\pm 112$ mins) and  9:30 am ($\pm 114$ mins)  respectively. For each window, multiple features from SC peaks, Skin Conductance Level (SCL), accelerometer, and temperature were computed. For SC Peak features, first peaks and artifacts were detected. Then, features listed in Table \ref{tab:wearable-feat} were computed for both, all detected peaks, and only non-artifact peaks.

    \item We collected metadata from mobile phones, including call, SMS, and screen usage information. These data encompassed timestamps, call/SMS duration, and counts of incoming/outgoing calls and SMS. We did not collect call/SMS content or details about specific applications for privacy reasons. We segmented the 24-hour day into three 7-hour windows and one 3-hour window. Finally, we extracted various features for both the entire day and these four-time windows, as elaborated in Table \ref{tab:phone-feat}.

    \item Weather: Multiple features about the weather are extracted from DarkSky’s weather API \cite{weather}. Features include temperature, sunlight,  barometric pressure, wind, and the difference between that day’s weather and a rolling average. Details are provided in Table \ref{tab:weather_des}.

    \item Surveys: Questionnaires were filled by participants every day about academic activities, exercise, extracurricular activities, nap count, interactions,  and caffeine and drug intake. 
\end{itemize}

\textit{Ground truth collection}: 
The following questions asked in the morning survey are used as ground truth sleep,
\begin{itemize}
    \item What time did you try to fall asleep? ($t_s$)
\item How long did it take you to fall asleep? ($t_1$)

\item What time did you finally wake up? ($t_2$)

\item How many times did you awaken?? ($t_3$)

\item How long were you awake for? ($t_4$)
\end{itemize}
Finally time in bed, $y$ was calculated as follows: $y = t_2 - t_s - t_1 - t_4$

\begin{table}[hb!]
\resizebox{\columnwidth}{!}{%
\begin{tabular}{|ll|}
\hline
\multicolumn{2}{|l|}{\textbf{Skin Conductance (SC) Peak Features}}                                                                                                                                                                                                    \\ \hline
\multicolumn{1}{|l|}{Skin Conductance Peak}                         & \begin{tabular}[c]{@{}l@{}}Peaks detected in EDA using EDA Explorer package\cite{EDA_Explorer} \end{tabular}                                              \\ \hline
\multicolumn{1}{|l|}{Sum AUC}                         & \begin{tabular}[c]{@{}l@{}}The sum of the Area Under Curve (AUC) of all peaks \end{tabular}                                              \\ \hline
\multicolumn{1}{|l|}{Sum   AUC Full}                  & Sum of AUC of peaks where the amplitude is   calculated as the difference from the base tonic level                                                                                                                                             \\ \hline
\multicolumn{1}{|l|}{Median   RiseTime}               & Median rise time of peaks (seconds)                                                                                                                                                                                  \\ \hline
\multicolumn{1}{|l|}{Median   Amplitude}              & Median amplitude of peaks (uS)                                                                                                                                                                                       \\ \hline
\multicolumn{1}{|l|}{Count   Peaks}                   & Number of peaks detected                                                                                                                                                                                             \\ \hline
\multicolumn{1}{|l|}{SD   Peaks 30 min}               & 
Standard deviation of \# of peaks per 30 minute epoch\\ \hline
\multicolumn{1}{|l|}{Med   Peaks 30 min}              &      Median of \# of peaks per 30 minute epoch                                                                                                 \\ \hline
\multicolumn{1}{|l|}{Percent   Med Peak}              & Percentage of the signal containing 1-minute epochs with greater than 5 peaks                                                                                                                                                                  \\ \hline                                             
\multicolumn{2}{|l|}{\textbf{Skin   Conductance Level (SCL) Features}}                                                                                                         \\ \hline
\multicolumn{1}{|l|}{Percent Off}                     & Percentage of period where the device was off                                                                                                                                                                            \\ \hline
\multicolumn{1}{|l|}{MaxUnnorm}                       & Maximum level of un normalized EDA signal                                                                                                                                                                            \\ \hline
\multicolumn{1}{|l|}{MedUnnorm}                       & Median of normalized EDA signal                                                                                                                                                                                      \\ \hline
\multicolumn{1}{|l|}{MeanUnnorm}                      & Mean of unnormalized EDA signal                                                                                                                                                                                      \\ \hline
\multicolumn{1}{|l|}{Median Norm}                     & Median of z score normalized EDA signal                                                                                                                                                                              \\ \hline
\multicolumn{1}{|l|}{SD Norm}                         & Standard Deviation of z score normalized EDA   signal                                                                                                                                                                \\ \hline
\multicolumn{1}{|l|}{Mean Deriv}                      & Mean derivative of z score normalized EDA   signal (uS/second)                                                                                                                                                       \\ \hline

\multicolumn{2}{|l|}{\textbf{Accelerometer Features}}                                                                                                                                                                                                                       \\ \hline
\multicolumn{1}{|l|}{Step Count}                      & Number of steps detected                                                                                                                                                                                             \\ \hline
\multicolumn{1}{|l|}{Mean Movement Step Time}         & \begin{tabular}[c]{@{}l@{}}Average number of samples (at 8Hz) between   two steps (aggregated first to 1 minute,\\  then we take the mean of only the   parts of this signal occurring during movement)\end{tabular} \\ \hline
\multicolumn{1}{|l|}{Stillness Percent}               & Percentage of time the person spent nearly   motionless                                                                                                                                                              \\ \hline
\multicolumn{1}{|l|}{Sum Stillness Weighted AUC}      & Weight the peak AUC signal by how still the   user was every 5 minutes and sum                                                                                                                                       \\ \hline
\multicolumn{1}{|l|}{Sum Steps Weighted AUC}          & Weight the peak AUC signal by the step count   over every 5 minutes and sum                                                                                                                                          \\ \hline
\multicolumn{1}{|l|}{Sum Stillness Weighted Peaks}    & Multiply the number of peaks every 5 minutes   by the amount of stillness during that period                                                                                                                         \\ \hline
\multicolumn{1}{|l|}{Max Stillness Weighted Peaks}    & The max value for the \# peaks * stillness   for any five minute period                                                                                                                                              \\ \hline
\multicolumn{1}{|l|}{Sum Steps Weighted Peaks}        & Divide the number of peaks every five minutes by   step count and sum                                                                                                                                                    \\ \hline
\multicolumn{1}{|l|}{Med Steps Weighted Peaks}        & Average value for the number of peaks/step count every 5 mins                                                                                                                                                    \\ \hline

\multicolumn{2}{|l|}{\textbf{Skin   Temperature (ST) Features}}                                                                                                                                                                                                              \\ \hline
                            
\multicolumn{1}{|l|}{Max Raw Temp}                    & The maximum of the raw temperature signal   (°C)                                                                                                                                                                     \\ \hline
\multicolumn{1}{|l|}{Min Raw Temp}                    & The minimum of the raw temperature signal   (°C)                                                                                                                                                                     \\ \hline
\multicolumn{1}{|l|}{SD Raw Temp}                     & The standard deviation of the raw   temperature signal                                                                                                                                                               \\ \hline
\multicolumn{1}{|l|}{Med Raw Temp}                    & Median of the raw   temperature                                                                                                                                                                                               \\ \hline
\multicolumn{1}{|l|}{Sum Temp Weighted AUC}           & Sum of peak AUC divided by the average temp   every 5 mins                                                                                                                                                           \\ \hline
\multicolumn{1}{|l|}{Sum Temp Weighted Peaks}         & Number of peaks divided by the average temp   every 5 mins                                                                                                                                                           \\ \hline
\multicolumn{1}{|l|}{Max Temp Weighted Peaks}         & The maximum number of peaks in any 5 minute   period divided by the average temp                                                                                                                                     \\ \hline
\multicolumn{1}{|l|}{SD Stillness Temp}               & Standard deviation of the temperature recorded during   periods when the person was still                                                                                                                                       \\ \hline
\multicolumn{1}{|l|}{Med   Stillness Temp}            & Median of the temperature when the person   was still                                                                                                                                                                \\ \hline
\end{tabular}}
\caption{Description of features extracted from wearable device}
\label{tab:wearable-feat}
\end{table}

\begin{table}[htb!]
\resizebox{0.9\columnwidth}{!}{%
\begin{tabular}{|ll|}
\hline
\multicolumn{2}{|l|}{\textbf{Phone Call Features}}                                                                                               \\ \hline
\multicolumn{1}{|l|}{Total Number}          & Total number of calls                                                                              \\ \hline
\multicolumn{1}{|l|}{Unique Number}         & Number of people connected with   (based on phone number)                                          \\ \hline
\multicolumn{1}{|l|}{Duration Statistics}   & For the duration of the call   compute Mean, Median, Standard deviation, and Total(Sum)             \\ \hline
\multicolumn{1}{|l|}{Timestamp Statistics}  & For the Timestamp when the call  started compute the Mean, Median, and Standard deviation           \\ \hline

\multicolumn{1}{|l|}{\textbf{SMS Features}} &                                                                                                    \\ \hline
\multicolumn{1}{|l|}{Total Number}          & Total number of SMS                                                                                \\ \hline
\multicolumn{1}{|l|}{Unique Number}         & Number of people connected with   (based on phone number)                                          \\ \hline
\multicolumn{1}{|l|}{Timestamp Statistics}  & For the Timestamp when SMS is   sent compute Mean, Median, Standard deviation                      \\ \hline

\multicolumn{2}{|l|}{\textbf{Screen   Usage Features}}                                                                                           \\ \hline
\multicolumn{1}{|l|}{Total Number}          & Total number of the screen on   instances                                                          \\ \hline
\multicolumn{1}{|l|}{Duration Statistics}   & For the duration of the screen   being on compute Mean, Median, Standard deviation, and Total (Sum) \\ \hline
\multicolumn{1}{|l|}{Timestamp Statistics}  & For the Timestamp when the screen   was turned on compute the Mean, Median, and Standard deviation  \\ \hline
\end{tabular}}
\caption{Description of features computed from Phone metadata}
\label{tab:phone-feat}
\end{table}

\begin{table}[htb!]
\resizebox{\columnwidth}{!}{%
\begin{tabular}{|ll}
\hline
\multicolumn{2}{|l}{\textbf{Weather   Features}}                                                                                                                                                                                                                                                        \\ \hline
\multicolumn{1}{|l|}{Sunrise}                                   & \multicolumn{1}{l|}{Sunrise time UTC}                                                                                                                                                                                                 \\ \hline
\multicolumn{1}{|l|}{Moon\_phase}                               & \multicolumn{1}{l|}{The moon phase value on a scale of $0-1$(new moon-full moon)}                                                                                                                                                     \\ \hline
\multicolumn{1}{|l|}{Apparent\_temp\_max}                       & \multicolumn{1}{l|}{Maximum apparent temperature of the day in Fahrenheit}                                                                                                                                                            \\ \hline
\multicolumn{1}{|l|}{Apparent\_temp\_min}                       & \multicolumn{1}{l|}{Minimum apparent temperature of the day in Fahrenheit}                                                                                                                                                            \\ \hline
\multicolumn{1}{|l|}{Temperature\_max}                          & \multicolumn{1}{l|}{Maximum temperature of the day in Fahrenheit}                                                                                                                                                                     \\ \hline
\multicolumn{1}{|l|}{Temperature\_min}                          & \multicolumn{1}{l|}{Minimum temperature of the day in Fahrenheit}                                                                                                                                                                     \\ \hline
\multicolumn{1}{|l|}{avg\_cloud\_cover}                         & \multicolumn{1}{l|}{Percentage of sky covered by cloud on a scale of 0-1}                                                                                                                                                             \\ \hline
\multicolumn{1}{|l|}{avg\_dew\_point}                           & \multicolumn{1}{l|}{Average dew point temperature}                                                                                                                                                                                    \\ \hline
\multicolumn{1}{|l|}{avg\_humidity}                             & \multicolumn{1}{l|}{Daily average value of humidity on a scale of 0-1}                                                                                                                                                                \\ \hline
\multicolumn{1}{|l|}{avg\_pressure}                             & \multicolumn{1}{l|}{Average atmospheric pressure on the sea level in hPa}                                                                                                                                                             \\ \hline
\multicolumn{1}{|l|}{Morning\_pressure\_change}                 & \multicolumn{1}{l|}{\begin{tabular}[c]{@{}l@{}}Trinary value of pressure difference between midnight and noon    (rising, falling, steady)\end{tabular}}                                                                            \\ \hline
\multicolumn{1}{|l|}{Evening\_pressure\_change}                 & \multicolumn{1}{l|}{\begin{tabular}[c]{@{}l@{}}Trinary value of pressure difference between noon and midnight  (rising, falling, steady)\end{tabular}}                                                                              \\ \hline
\multicolumn{1}{|l|}{avg\_visibility}                           & \multicolumn{1}{l|}{Average visibility in meters}                                                                                                                                                                                     \\ \hline
\multicolumn{1}{|l|}{weather\_precip\_probability}              & \multicolumn{1}{l|}{Precipitation probability}                                                                                                                                                                                        \\ \hline
\multicolumn{1}{|l|}{Temperature\_rolling\_mean}                & \multicolumn{1}{l|}{Rolling average of temperature}                                                                                                                                                                                   \\ \hline
\multicolumn{1}{|l|}{Temperature\_rolling\_std}                 & \multicolumn{1}{l|}{Rolling standard deviation in temperature}                                                                                                                                                                        \\ \hline
\multicolumn{1}{|l|}{Temperature\_today\_vs\_avg\_past}         & \multicolumn{1}{l|}{Difference between today's temperature and rolling average}                                                                                                                                                       \\ \hline
\multicolumn{1}{|l|}{apparentTemperature\_rolling\_mean}        & \multicolumn{1}{l|}{Rolling average of  apparent   temperature}                                                                                                                                                                       \\ \hline
\multicolumn{1}{|l|}{apparentTemperature\_rolling\_std}         & \multicolumn{1}{l|}{Rolling standard deviation in apparent temperature}                                                                                                                                                               \\ \hline
\multicolumn{1}{|l|}{apparentTemperature\_today\_vs\_avg\_past} & \multicolumn{1}{l|}{Difference in today's apparent temperature and its rolling average}                                                                                                                                               \\ \hline
\multicolumn{1}{|l|}{pressure\_rolling\_mean}                   & \multicolumn{1}{l|}{Rolling average of pressure}                                                                                                                                                                                      \\ \hline
\multicolumn{1}{|l|}{pressure\_rolling\_std}                    & \multicolumn{1}{l|}{Rolling standard deviation of pressure}                                                                                                                                                                           \\ \hline
\multicolumn{1}{|l|}{pressure\_today\_vs\_avg\_past}            & \multicolumn{1}{l|}{Difference between today's pressure and its rolling average}                                                                                                                                                      \\ \hline
\multicolumn{1}{|l|}{cloudCover\_rolling\_mean}                 & \multicolumn{1}{l|}{Rolling average of cloud cover}                                                                                                                                                                                   \\ \hline
\multicolumn{1}{|l|}{cloudCover\_rolling\_std}                  & \multicolumn{1}{l|}{Rolling standard deviation in cloud cover}                                                                                                                                                                        \\ \hline
\multicolumn{1}{|l|}{cloudCover\_today\_vs\_avg\_past}          & \multicolumn{1}{l|}{Difference between today's cloud cover and its rolling average}                                                                                                                                                   \\ \hline
\multicolumn{1}{|l|}{humidity\_rolling\_mean}                   & \multicolumn{1}{l|}{Rolling average of humidity}                                                                                                                                                                                      \\ \hline
\multicolumn{1}{|l|}{humidity\_rolling\_std}                    & \multicolumn{1}{l|}{Rolling standard deviation in humidity}                                                                                                                                                                           \\ \hline
\multicolumn{1}{|l|}{humidity\_today\_vs\_avg\_past}            & \multicolumn{1}{l|}{Difference between today's humidity and its rolling average}                                                                                                                                                      \\ \hline
\multicolumn{1}{|l|}{windSpeed\_rolling\_mean}                  & \multicolumn{1}{l|}{Rolling average of wind speed}                                                                                                                                                                                    \\ \hline
\multicolumn{1}{|l|}{windSpeed\_rolling\_std}                   & \multicolumn{1}{l|}{Rolling standard deviation in wind speed}                                                                                                                                                                         \\ \hline
\multicolumn{1}{|l|}{windSpeed\_today\_vs\_avg\_past}           & \multicolumn{1}{l|}{Difference between today's wind speed and its rolling average}                                                                                                                                                    \\ \hline
\multicolumn{1}{|l|}{precipProbability\_rolling\_mean}          & \multicolumn{1}{l|}{Rolling average of precipitation probability}                                                                                                                                                                     \\ \hline
\multicolumn{1}{|l|}{precipProbability\_rolling\_std}           & \multicolumn{1}{l|}{Rolling standard deviation in precipitation probability}                                                                                                                                                          \\ \hline
\multicolumn{1}{|l|}{precipProbability\_today\_vs\_avg\_past}   & \multicolumn{1}{l|}{Difference between current precipitation probability and its rolling   average}                                                                                                                                   \\ \hline
\multicolumn{1}{|l|}{sunlight}                                  & \multicolumn{1}{l|}{Duration of sunlight in seconds}                                                                                                                                                                                  \\ \hline
\multicolumn{1}{|l|}{quality\_of\_day}                          & \multicolumn{1}{l|}{\begin{tabular}[c]{@{}l@{}}Quality of the day define in terms of 8 categories in the range $\{-4,4\}$:\\ clear= 4, partly-cloudy= 3, cloudy= 2,  wind=1, fog= -1,  rain= -2,  sleet= -3,   snow= -4\end{tabular}} \\ \hline
\multicolumn{1}{|l|}{avg\_quality\_of\_day}                     & \multicolumn{1}{l|}{Average value for quality\_of\_day}                                                                                                                                                                               \\ \hline
\multicolumn{1}{|l|}{precipType}                                & \multicolumn{1}{l|}{\begin{tabular}[c]{@{}l@{}}Type of precipitation as integer: None = 0, Rain = 1, Hail = 2, Sleet = 3, Snow = 4, Other = 5\end{tabular}}                                                                        \\ \hline
\multicolumn{1}{|l|}{max\_precip\_intensity}                    & \multicolumn{1}{l|}{Maximum Precipitation volume in mm}                                                                                                                                                                               \\ \hline
\multicolumn{1}{|l|}{median\_wind\_speed}                       & \multicolumn{1}{l|}{Median wind speed of the day in meter/sec}                                                                                                                                                                        \\ \hline
\multicolumn{1}{|l|}{median\_wind\_bearing}                     & \multicolumn{1}{l|}{Median wind bearing of the day in degrees}            \\ \hline                                                                                                                               
\end{tabular}}
\caption{Description of weather features}
\label{tab:weather_des}
\end{table}

\subsection{Network Characteristics}\label{supp-net}

\subsubsection{Networks from survey information}
Participants ranked their connections (e.g., friends, roommates) in pre- and post-study surveys, assigning up to three people per question with decreasing points from 3 to 1 based on closeness. These rankings were aggregated into scores to determine edge strengths in graphs, with separate graphs created for study participants based on both surveys.

 Many participants mentioned out-of-study people; we created separate pre-study and post-study graphs for in-study participants. When creating graphs between only-study participants, the percentage (rounded to the nearest integer) of unique connections that were retained from the network of all connections in the pre-study survey from cohort 1 to 7 was $18\%,36\%,17\%,15\%,21\%,10\%$ and $14\%$ respectively. For the post-study survey, the proportion retained from cohorts 1 to 7 was $15\%,34\%,13\%,17\%,20\%,10\%$, and $16\%$ respectively. Degree distributions in graphs with all participants were skewed due to out-of-study participants. Filtering to only-study participants provided a more balanced view, with most cohorts having a mean of approximately 2 nodes as shown in Table \ref{tab:deg_survey_all}.

\begin{figure}
    \centering
    \includegraphics[width=0.6\textwidth]{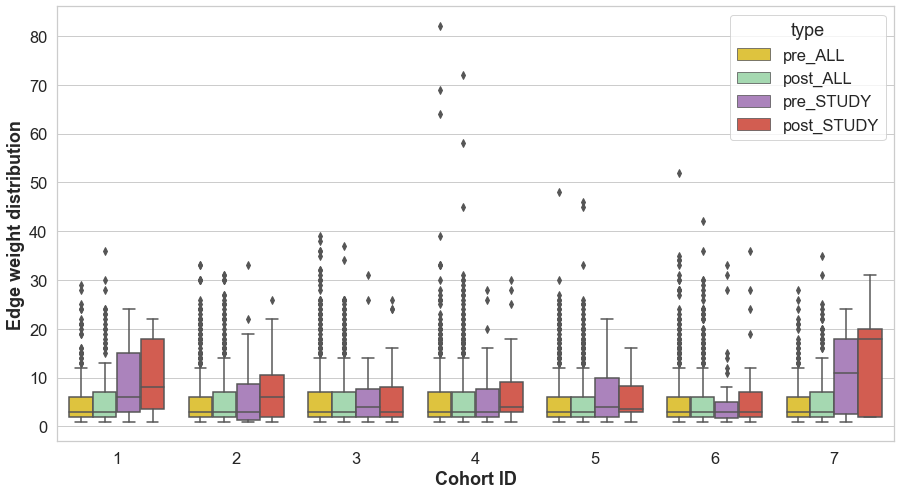}
    \caption{Distribution of the strength of edges in networks obtained from social survey data. The legend pre\_All and post\_ALL indicate the network created from pre and post\_surveys with all participants indicated by the user. pre\_study and post\_study refer to networks created between study participants only.}
    \label{fig:edge_all}
\end{figure}

To assess connection strength, we plotted edge strength distributions in Fig. \ref{fig:edge_all}. The boxplots revealed large interquartile ranges within cohorts, suggesting diverse connection strengths. Out-of-study participants caused outliers in overall graphs, while in-cohort graphs showed notable variability, indicating closer relationships among study participants, confirming their communication and acquaintance as required for the study.

\begin{table}[htb!]
\centering
\resizebox{0.7\textwidth}{!}{%
\begin{tabular}{|l|}
\hline
1) Name your  roommates/flatmates? (if any) \\ \hline
2) If a natural disaster/tragedy affects   Boston, who would you call first? second? \\ \hline
3) Who do you talk to about personal   matters (love life, concerns, family matters, etc)? \\ \hline
4) Who do you talk to about   work/research/classes? \\ \hline
5) Who do you talk to about friends and   other people you know? \\ \hline
6) Who do you talk to about   media/entertainment (sports, movies, tech gadgets, music, video games)? \\ \hline
7) If you have to move to a new   apartment/house, who would you ask for help? \\ \hline
8) Who do you study with? \\ \hline
9) Who do you go shop, party or play with   (including video games)? \\ \hline
10) Who do you share ideas with? \\ \hline
11) Who do you often disagree with (real   person, not media character)? \\ \hline
12) Who spends the most time at your   apartment/dorm excluding the people you live with? \\ \hline
13) Do you hang out at someone else's   apartment/dorm? If yes, whose? \\ \hline
\end{tabular}%
}
\caption{List of questions in pre-study and post-study surveys}
\label{tab:social_question}
\end{table}

\begin{table}[htb!]
\centering
\resizebox{0.7\textwidth}{!}{%
\begin{tabular}{|l|l|l|l|l|l|l|l|l|}
\hline
\textbf{Cohort} & \textbf{Survey type} & \textbf{Mean} & \textbf{Median} & \textbf{\begin{tabular}[c]{@{}l@{}}Standard\\  Deviation\end{tabular}}   & \textbf{Minimum}    & \textbf{\begin{tabular}[c]{@{}l@{}}25th\\  Percentile\end{tabular}} & \textbf{\begin{tabular}[c]{@{}l@{}}75th \\ Percentile\end{tabular}} & \textbf{Maximum}  \\ \hline
\multirow{2}{*}{1} 
 & pre\_STUDY & 1.22 & 1 & 0.43 & 1 & 1 & 1 & 2 \\ \cline{2-9} 
 & post\_STUDY & 1.29 & 1 & 0.47 & 1 & 1 & 2 & 2 \\ \hline
\multirow{2}{*}{2} 
 & pre\_STUDY & 1.56 & 1 & 1.42 & 1 & 1 & 1.75 & 7 \\ \cline{2-9} 
 & post\_STUDY & 1.41 & 1 & 1.28 & 1 & 1 & 1 & 6 \\ \hline
\multirow{2}{*}{3} 
 & pre\_STUDY & 1.62 & 1 & 0.94 & 1 & 1 & 2 & 4 \\ \cline{2-9} 
 & post\_STUDY & 1.86 & 1 & 1.23 & 1 & 1 & 2.5 & 7 \\ \hline
\multirow{2}{*}{4} 
 & pre\_STUDY & 2.12 & 2 & 1.18 & 1 & 1 & 3 & 5 \\ \cline{2-9} 
 & post\_STUDY & 1.76 & 1 & 1.17 & 1 & 1 & 2 & 5 \\ \hline
\multirow{2}{*}{5} 
 & pre\_STUDY & 1.52 & 1 & 0.91 & 1 & 1 & 2 & 5 \\ \cline{2-9} 
 & post\_STUDY & 1.83 & 2 & 0.87 & 1 & 1 & 3 & 3 \\ \hline
\multirow{2}{*}{6} 
 & pre\_STUDY & 2.21 & 2 & 1.41 & 1 & 1 & 3 & 6 \\ \cline{2-9} 
 & post\_STUDY & 2 & 2 & 1.07 & 1 & 1 & 3 & 4 \\ \hline
\multirow{2}{*}{7} 
 & pre\_STUDY & 1.47 & 1 & 0.74 & 1 & 1 & 2 & 3 \\ \cline{2-9} 
 & post\_STUDY & 1.5 & 1 & 0.8 & 1 & 1 & 2 & 3 \\ \hline
\end{tabular}%
}
\caption{Distribution of the number of connections/degree in networks obtained from social survey data. The survey type pre\_study and post\_study refer to pre-study and post-study networks created between study participants only.}
\label{tab:deg_survey_all}
\end{table}

\subsubsection{Networks from Phone Data}
\begin{figure}[]
    \centering
    \includegraphics[width=0.55\textwidth]{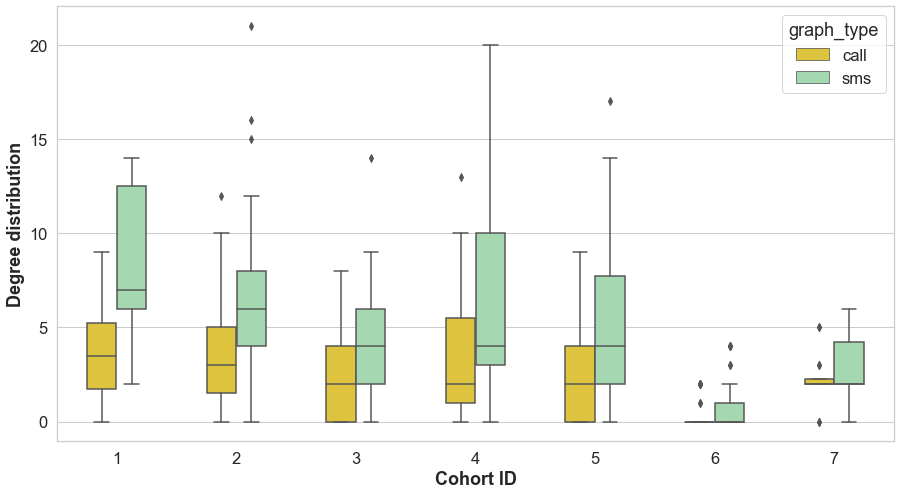}
    \caption{Distribution of degree in graphs created from phone data in each cohort}
    \label{fig:phone_degree}
\end{figure}

\begin{figure}[]
	\centering
	\subfloat[][\centering Distribution of edge strength(number of SMS exchanged) in SMS graphs in each cohort]{{\includegraphics[width = 0.37\textwidth]{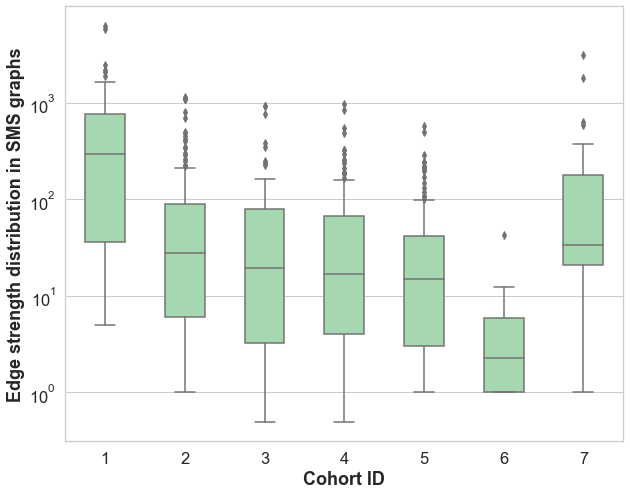} }}%
	\qquad
	\subfloat[][\centering Distribution of edge strength(aggregated call duration in seconds) in call graphs in each cohort]{{\includegraphics[width = 0.37\textwidth]{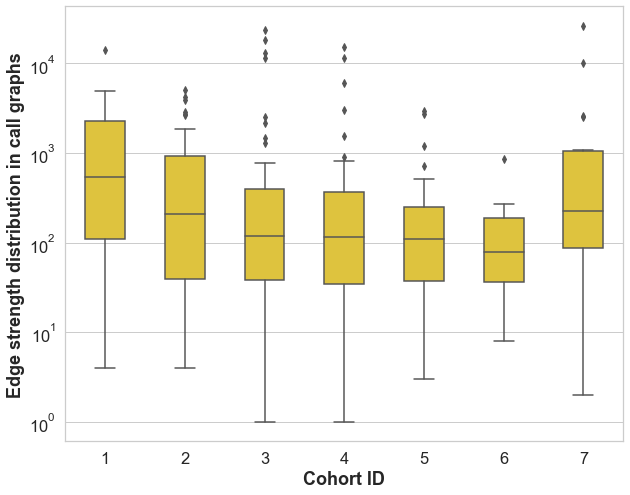} }}%
	\caption{Distribution of edge strength (phone activity between users) in networks created from phone metadata}%
	\label{fig:edge_phone_graph}%
\end{figure}

We construct graphs from metadata collected from phones for each cohort. To understand how many users a node was connected to, the degree density for each cohort is presented in Fig. \ref{fig:phone_degree}. For further understanding of the network characteristics within a cohort, the distribution of degree and edge strength is provided for SMS and call graphs in Fig. \ref{fig:edge_phone_graph}(a) and Fig. \ref{fig:edge_phone_graph}(b)  respectively.

\end{document}